\documentclass[10pt,twocolumn,letterpaper]{article}
\usepackage[pagenumbers]{wacv}

\usepackage{float}
\usepackage{enumitem}

\usepackage{amsmath}
\usepackage{amssymb}
\usepackage{amsfonts}
\usepackage{bm}

\usepackage{algorithm}
\usepackage{algpseudocode}
\algrenewcomment[1]{\hfill$\triangleright$~#1}
\algrenewcommand\algorithmicrequire{\textbf{Input:}}
\algrenewcommand\algorithmicensure{\textbf{Return:}}

\usepackage{multirow}
\usepackage{booktabs}
\usepackage{graphicx}
\usepackage{subcaption}
\usepackage{arydshln}

\usepackage{wrapfig}

\usepackage[table, dvipsnames]{xcolor}

\usepackage{thmtools}
\usepackage{thm-restate}

\usepackage{titletoc}

\usepackage[most]{tcolorbox}

\definecolor{DarkGreen}{rgb}{0.00, 0.40, 0.00}
\definecolor{RoyalBlue}{rgb}{0.25, 0.35, 0.74}
\definecolor{Crimson}{rgb}{0.86, 0.08, 0.24}
\definecolor{ChromeYellow}{rgb}{1.0, 0.65, 0.0}

\usepackage{pifont}
\newcommand{\xmark}{\ding{55}}
\newcommand{\method}{V2C-Edit}
\newcommand{\argmax}{\operatornamewithlimits{argmax}}
\newcommand{\first}[1]{\textbf{#1}}
\newcommand{\second}[1]{\underline{#1}}
\colorlet{ourshl}{blue!10}

\usepackage[most]{tcolorbox}

\newtcolorbox{highlightbox}{
    colback=blue!3,
    colframe=blue!40,
    boxrule=0.7pt,
    arc=3pt,
    left=4pt,
    right=4pt,
    top=3pt,
    bottom=3pt,
    breakable,
    before skip=10pt,
    after skip=10pt
}

\lstdefinestyle{promptbox}{
    backgroundcolor=\color{gray!10},
    basicstyle=\ttfamily\small,
    breaklines=true,
    breakatwhitespace=true,
    frame=single,
    framerule=0.5pt,
    rulecolor=\color{gray!50},
    columns=fullflexible,
    xleftmargin=5pt,
    xrightmargin=5pt,
    aboveskip=8pt,
    belowskip=8pt,
}

\usepackage{xcolor}

\newcolumntype{+}{>{\global\let\currentrowstyle\relax}}
\newcolumntype{^}{>{\currentrowstyle}}

\definecolor{noncomparablegray}{RGB}{145,145,145}

\usepackage{setspace}

\usepackage[most]{tcolorbox}
\tcbuselibrary{listings,breakable}
\newtcblisting{promptbox}[1][]{
    listing only,
    title={#1},
    fonttitle=\bfseries,
    listing options={
        basicstyle=\ttfamily\small,
        breaklines=true,
        breakindent=0pt,
        breakautoindent=false,
        columns=fullflexible
    }
}

\usepackage{tocloft}
\usepackage{titletoc}
\newcommand\DoToC{%
  \startcontents
  {\bfseries \noindent Table of Contents\par}
  \vskip5pt
  \printcontents{}{1}{\hrule\vskip5pt}
  \vskip5pt\hrule\vskip5pt
}

\definecolor{wacvblue}{rgb}{0.21,0.49,0.74}
\usepackage[pagebackref,breaklinks,colorlinks,allcolors=wacvblue]{hyperref}

\title{VLM as an Editing Pilot: VLM-Steered Velocity Control for Flow-Based \\ Image Editing}

\author{
Kuan-Shen Chiu\textsuperscript{1,4 *}
\qquad
Jia-Wei Liao\textsuperscript{1}\thanks{Equal contribution.}
\qquad
Yu-Chieh Chao\textsuperscript{3}
\qquad
Li-Xuan Peng\textsuperscript{2} \\
Wei-Yun Ma\textsuperscript{4}\thanks{Corresponding author.}
\qquad
Cheng-Fu Chou\textsuperscript{1}
\vspace{0.5em} \\
\textsuperscript{1}National Taiwan University \quad
\textsuperscript{2}National Tsinghua University \quad
\textsuperscript{3}University of California, Davis \\
\textsuperscript{4}Institute of Information Science, Academia Sinica
}

\begin{document}
\maketitle
\begin{abstract}
Text-guided image editing requires realizing the requested semantic change while preserving content unrelated to the instruction. FlowEdit provides an efficient inversion-free solution, but its source-anchored coupling can become overly restrictive, causing \emph{zero-edit}, where the output remains close to the source and the requested change fails to emerge. We find that source coupling directly governs the preservation--editing trade-off, while a fixed coupling schedule cannot adapt to timestep-specific editing needs. Motivated by this observation, we propose \method{}, a training-free VLM-steered velocity control framework. \method{} constructs complementary coupling-controlled velocities and dynamically selects among them using Instruction-Level Similarity (ILS), which evaluates whether the predicted transformation follows the editing instruction. Adaptive spatial masking further preserves regions unrelated to the edit. Experiments on MagicBrush and PIE-Bench with FLUX and Stable Diffusion 3 show that \method{} consistently improves instruction alignment over FlowEdit while maintaining competitive structural preservation, and that ILS better reflects editing effectiveness than target-caption CLIP similarity.
Our code is available at \url{https://github.com/Kuanshen77/V2C-Edit}.
\end{abstract}

\section{Introduction}

Text-guided image editing aims to modify an input image according to a user instruction while preserving content unrelated to the requested change. This requires a careful balance between semantic modification and source preservation: the edit must be strong enough to realize the instruction, yet localized enough to avoid unnecessary changes to the original image. Achieving both properties is therefore a central challenge in practical image editing.

The success of diffusion models~\cite{rombach2022high} and flow models~\cite{sd3medium, flux, lipman2023flow, liu2023flow, esser2024scaling} has enabled a range of training-free editing methods,  including inversion-based methods~\cite{song2021denoising, huberman2024edit, mokady2023null, ju2024pnp, wallace2023edict, wang2025rfedit, miyake2025negative, huang2025dual, han2024proxedit, pan2023effective, xu2025unveil, nguyen2025h}, internal representation manipulation methods~\cite{hertz2023prompttoprompt, cao2023masactrl, tumanyan2023plug, avrahami2025stable, kim2025reflex, bai2025edicho, liao2026freqedit, shen2025qk, liu2024towards}, and inversion-free approaches~\cite{meng2022sdedit, xu2023infedit, yoon2026splitflow, patel2025flowchef, lu2026chordedit, zhu2025kv}. FlowEdit~\cite{kulikov2025flowedit} further advances inversion-free editing by directly coupling source and target references along the editing trajectory, providing source preservation and improved model compatibility. However, our analysis reveals a fundamental preservation--editing trade-off: excessive source coupling can suppress the requested semantic change and produce \emph{zero-edit}, where the output remains nearly identical to the source image. Moreover, FlowEdit controls this behavior using a single fixed coupling switch point, which cannot adapt to different editing requirements across timesteps. We further observe that target-caption CLIP similarity is an unreliable signal for such adaptation, since it can remain high even when the requested edit has not occurred.

To address these limitations, we propose \method{}, a VLM-steered closed-loop velocity controller for flow-based image editing. Our key idea is to replace the fixed coupling schedule with per-step semantic control. We first construct a compact set of coupling-controlled velocities that provide complementary preservation--editing behaviors, ranging from source-anchored to locally and persistently decoupled alternatives. We then introduce \emph{Instruction-Level Similarity} (ILS), which uses a VLM to assess whether the predicted transformation of each candidate follows the editing instruction. At each timestep, \method{} selects the velocity with the highest instruction-level score and progressively activates spatial masking once the requested semantics have sufficiently emerged, enabling the trajectory to adapt both \emph{how} and \emph{where} the image is edited.

We evaluate \method{} on MagicBrush~\cite{zhang2023magicbrush} and PIE-Bench~\cite{ju2024pnp} using FLUX~\cite{flux} and Stable Diffusion 3~\cite{sd3medium}. Across both backbones, \method{} consistently improves instruction alignment over FlowEdit while maintaining competitive structural preservation and substantially reducing zero-edit failures. Our analysis further shows that instruction-level VLM evaluation better reflects editing effectiveness than target-caption CLIP similarity, supporting its use as a semantic control signal.

Our contributions are summarized as follows:
\begin{enumerate}
\item We identify and analyze the \emph{zero-edit} failure mode in FlowEdit, showing that source coupling governs the preservation--editing trade-off and that a single fixed coupling switch point is insufficient for timestep-adaptive editing.
\item We propose \method{}, a training-free closed-loop velocity control framework that dynamically selects among complementary coupling-controlled velocities and applies adaptive spatial masking for source preservation.
\item We introduce Instruction-Level Similarity (ILS), a VLM-based semantic objective that evaluates whether the realized transformation follows the editing instruction and provides a more suitable control signal than target-caption CLIP similarity.
\item Extensive experiments on MagicBrush and PIE-Bench with FLUX and Stable Diffusion 3 demonstrate improved instruction alignment, reduced zero-edit failures, and competitive structural preservation.
\end{enumerate}

\section{Related Work}

Text-guided image editing aims to modify an image according to a text instruction while preserving content unrelated to the requested change. Existing training-free approaches can be broadly grouped into optimization-based editing, inversion-based editing, feature-based structure preservation, and inversion-free editing.

\paragraph{Optimization-based Editing.}
Optimization-based methods directly optimize an image, latent representation, or model input using guidance derived from a pre-trained generative model~\cite{hertz2023delta, kawar2023imagic, kim2024dreamsampler, koo2024posterior}. Most existing approaches are developed for diffusion models, while iRFDS~\cite{yang2025text_irfds} extends this paradigm to rectified flow models. These methods offer flexible control over the editing objective, but typically require iterative optimization for each input image, increasing inference-time computation.

\paragraph{Inversion-based Editing.}
A large family of zero-shot editing methods first invert the source image into a noisy latent or noise representation and then regenerate it under the target prompt~\cite{song2021denoising, huberman2024edit, mokady2023null, ju2024pnp}. Recent methods improve this process by reducing inversion errors or drift. RF Inversion~\cite{rout2025semantic} introduces a stochastic and optimal-control-based formulation for rectified flows, while DNAEdit~\cite{xie2025dnaedit} refines the recovered noise to reduce accumulated errors. Although these methods improve reconstruction and editing quality, their performance remains dependent on the quality of the inversion and subsequent prompt-conditioned regeneration.

\paragraph{Feature-based Structure Preservation.}
Built upon inversion-based editing pipelines, another line of work preserves source content by reusing or manipulating model-internal representations during editing. Prompt-to-Prompt~\cite{hertz2023prompttoprompt} controls spatial correspondence through cross-attention maps, while MasaCtrl~\cite{cao2023masactrl} leverages mutual self-attention to preserve local appearance and structure. Related approaches reuse attention maps or intermediate features to improve source fidelity~\cite{tumanyan2023plug, parmar2023zero, duan2024tuning}, with similar ideas later extended to rectified flow models~\cite{wang2025rfedit}. Such methods can provide strong structural preservation, but often depend on particular model architectures, attention modules, or internal representations.

\paragraph{Inversion-Free Editing.}
FlowEdit~\cite{kulikov2025flowedit} introduces an inversion-free alternative by constructing an editing trajectory directly between source and target distributions, without test-time optimization or manipulation of model internals. This design provides strong source preservation and improved model compatibility. However, the fixed coupling between the editing trajectory and the source image can make some edits overly conservative, leaving the requested semantic change insufficiently expressed. Our work focuses on this under-explored trade-off between source preservation and semantic modification in inversion-free flow editing. Instead of using a fixed coupling strategy, we dynamically steer the editing process according to instruction-level semantic feedback, allowing the trajectory to better adapt to the editing requirement while retaining source content.

\begin{figure*}
    \centering
    \includegraphics[width=\linewidth]{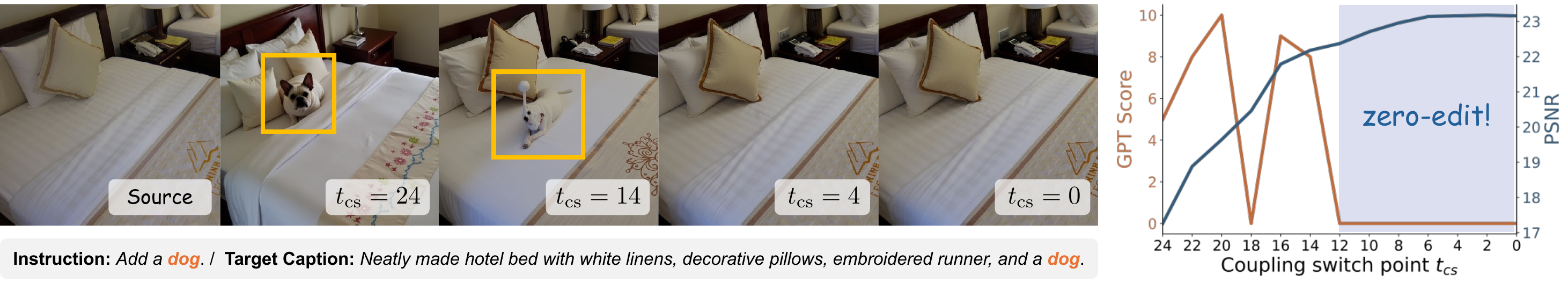}
    \caption{\textbf{Coupling controls the preservation-editing trade-off in FlowEdit.} As $t_{\mathrm{cs}}$ decreases, the coupled stage is retained for longer, increasing PSNR to the source image over the annotated unedited region but weakening the requested semantic change. In the example of \emph{``Add a dog,''} strong coupling eventually leads to \emph{zero-edit}, where the output remains nearly unchanged despite high preservation.}
    \label{fig:coupling_switch_point}
\end{figure*}

\section{Background of FlowEdit}

FlowEdit~\cite{kulikov2025flowedit} performs inversion-free editing in two stages. Given a source image $X^{\mathrm{src}}$ and source/target conditions $\mathcal{C}^{\mathrm{src}}$ and $\mathcal{C}^{\mathrm{tar}}$, it first performs coupled flow editing from the initial timestep to a \emph{coupling switch point} $t_{\mathrm{cs}}$, and then continues with standard target-conditioned flow sampling until $t=0$. Standard FlowEdit uses $t_{\mathrm{cs}} = 0$, maintaining the coupled construction throughout the editing trajectory.

\paragraph{Stage 1: Coupled Flow Editing.}
At timestep $t$, FlowEdit constructs the source and target references as
\begin{equation}
\begin{aligned}
    X_t^{\mathrm{src}} &= (1 - t) X_0^{\mathrm{src}} + t \epsilon_t,
    \quad \epsilon_t \sim \mathcal{N}(\mathbf{0}, \mathbf{I}), \\
    X_t^{\mathrm{tar}}
    & = X_t^{\mathrm{src}} + X_t^{\mathrm{edit}}-X^{\mathrm{src}},
    \label{eq:flowedit_xtar}
\end{aligned}
\end{equation}
which explicitly couples the target reference to the source reference, keeping the target trajectory anchored to the current editing state relative to the source. The editing trajectory is driven by the difference between the target- and source-conditioned velocity fields:
\begin{equation*}
\begin{aligned}
    V_t^\mathrm{edit} &= V_\theta(X_t^\mathrm{tar}, t, \mathcal{C}^\mathrm{tar}) - V_\theta(X_t^\mathrm{src}, t, \mathcal{C}^\mathrm{src}), \\
    X_{t - \Delta t}^\mathrm{edit} &= X_t^\mathrm{edit} - V_t^\mathrm{edit} \Delta t.
\end{aligned}
\end{equation*}

\paragraph{Stage 2: Target Flow Sampling.}
At the switching timestep, FlowEdit initializes $X_t^{\mathrm{tar}}$ using the same construction above and then discards the source branch. The remaining trajectory follows only the target-conditioned flow:
\begin{equation*}
    X_{t - \Delta t}^{\mathrm{tar}} = X_t^{\mathrm{tar}} - V_\theta(X_t^{\mathrm{tar}}, t, \mathcal{C}^{\mathrm{tar}}) \Delta t.
\end{equation*}

\section{Systematic Analysis of Flow-Based Editing}

\subsection{Rethinking Evaluation for Image Editing}
\begin{highlightbox}
\textbf{Finding 1.} Target-caption CLIP similarity can overlook edit-critical words, especially in long captions, and therefore does not reliably measure whether the requested edit occurs.
\end{highlightbox}

\begin{figure}
    \centering
    \includegraphics[width=0.9\linewidth]{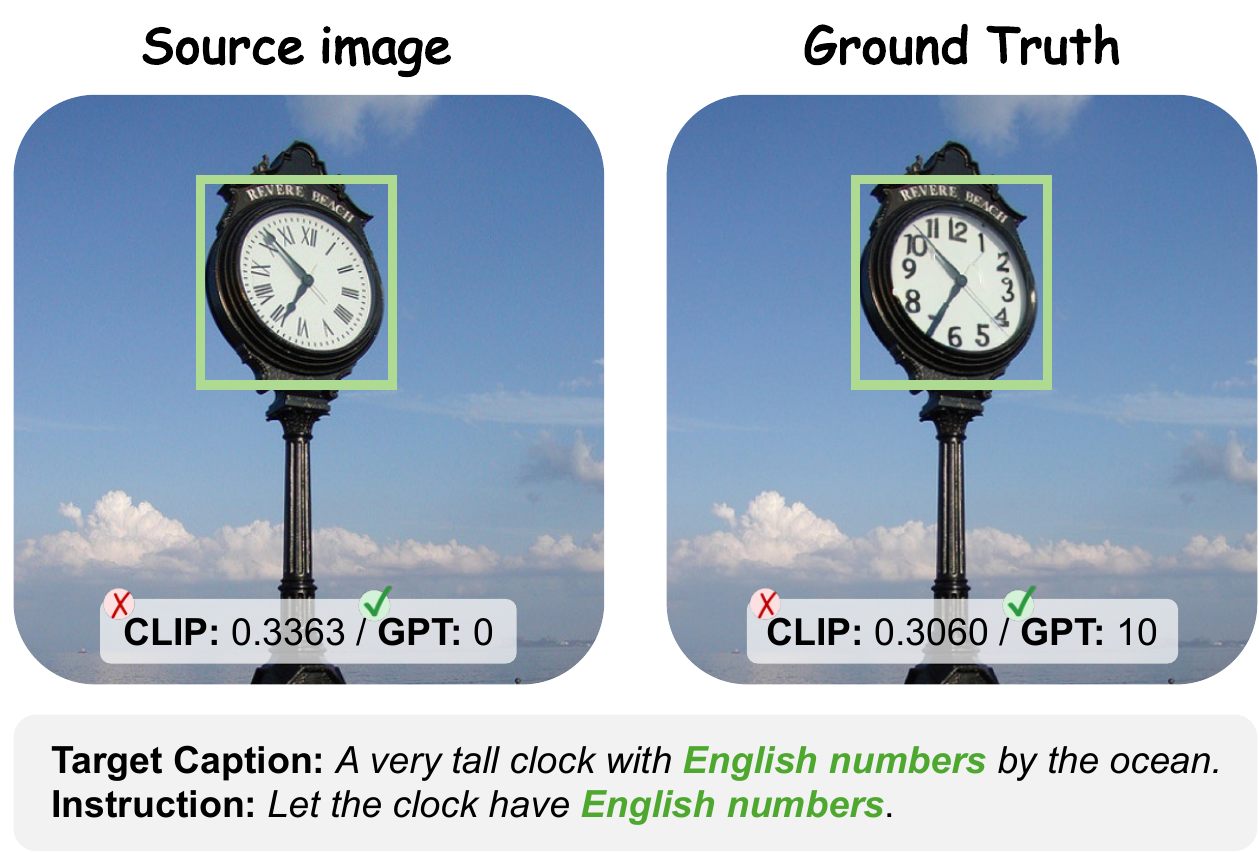}
    \caption{\textbf{Failure of target-caption CLIP similarity for evaluating image editing.} Although the ground-truth edit correctly changes the clock face to English numbers, it receives a lower CLIP similarity than the unedited source image. In contrast, the instruction-level GPT score correctly favors the edited result, showing that target-caption similarity may fail to reflect whether the requested edit is actually realized.}
    \label{fig:clip_failure}
\end{figure}

CLIP similarity~\cite{radford2021clip} is commonly used to assess semantic alignment between an edited image and the target caption. However, it evaluates only the final image-caption correspondence and does not explicitly measure whether the requested transformation from the source image has occurred. Moreover, because the caption is compressed into a global representation, edit-critical words can be underweighted, particularly in longer captions. As shown in Fig.~\ref{fig:clip_failure}, this can cause an unedited source image to score higher than a correctly edited result when most source semantics already match the target caption.

\begin{highlightbox}
\textbf{Finding 2.} Instruction-level VLM evaluation better reflects editing effectiveness than target-caption CLIP similarity.
\end{highlightbox}

This limitation motivates evaluating the edit directly against the instruction rather than relying on target-caption similarity. While humans can judge whether the observed transformation follows the requested instruction, repeated human evaluation is impractical. We therefore use a GPT evaluator~\cite{openaigpt41} to score the alignment between the edited image and the editing instruction, with details provided in Sec. \ref{sec:evaluation_metric}. As shown in Fig.~\ref{fig:clip_failure}, this GPT score better reflects editing effectiveness than target-caption 
CLIP similarity.

\begin{figure*}[t]
    \centering
    \includegraphics[width=\textwidth]{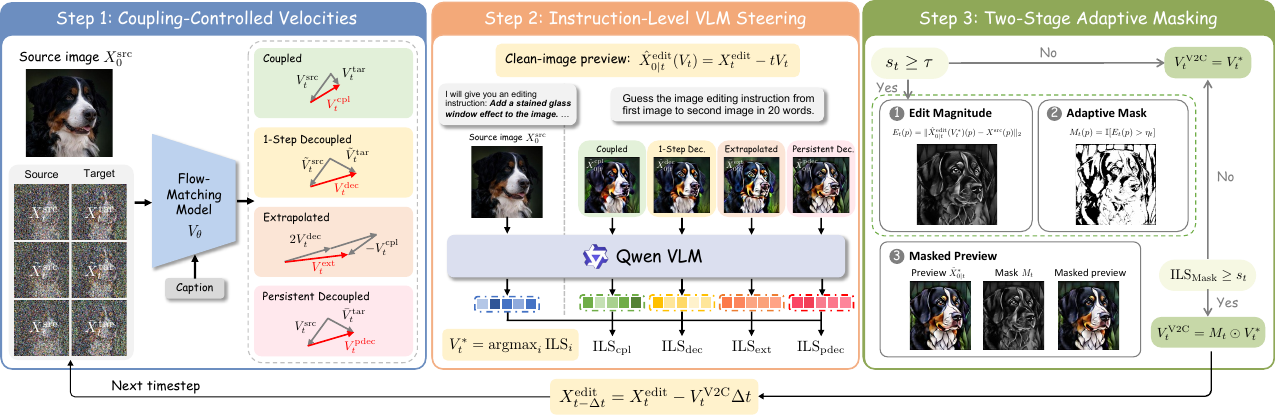}
    \caption{\textbf{Overview of the proposed V2C-Edit pipeline.} (1) Coupling-Controlled Velocities: Construct four editing velocities with different coupling behaviors. (2) Instruction-Level VLM Steering: Select the optimal velocity at each timestep using ILS. (3) Two-Stage Adaptive Masking: Apply spatial masking when semantic alignment is sufficient and the masked update maintains ILS. The VLM query templates are provided in Appendix ~\ref{app:prompt_tmp}.}
    
    \label{fig:pipeline}
\end{figure*}

\subsection{Coupling Analysis in Flow-based Editing}
\label{sec:coupling_analysis}
\begin{highlightbox}
\textbf{Finding 3.} Source coupling governs the preservation-editing trade-off, but a single fixed coupling switch point cannot adapt to timestep-specific editing needs.
\end{highlightbox}

FlowEdit~\cite{kulikov2025flowedit} anchors the target reference to the source through Eq.~\eqref{eq:flowedit_xtar}, while the coupling switch point $t_{\mathrm{cs}}$ determines how long this source-anchored construction is retained. As shown in Fig.~\ref{fig:coupling_switch_point}, decreasing $t_{\mathrm{cs}}$ improves preservation but weakens the requested semantic change; at the extreme, standard FlowEdit with $t_{\mathrm{cs}}=0$ produces \emph{zero-edit}, where the output remains nearly unchanged under the instruction. Increasing $t_{\mathrm{cs}}$ relaxes the coupling and enables the edit to emerge, but progressively reduces preservation in unedited regions. This reveals that coupling directly controls the preservation-editing trade-off. However, a single fixed $t_{\mathrm{cs}}$ imposes one coupling schedule over the entire trajectory, even though the required degree of coupling can vary across timesteps. This motivates replacing the fixed switch with per-step velocity control that dynamically adapts the coupling behavior according to the current editing requirement.

\section{\method{}: VLM-Steered Velocity Control for Image Editing}

Our analysis shows that inversion-free editing methods~\cite{kulikov2025flowedit} cannot consistently balance semantic editing and source preservation across different instructions and timesteps. To adapt this trade-off during generation, we propose \method{}, formulating image editing as a closed-loop velocity control problem. We first construct a set of coupling-controlled velocities that provide complementary preservation-editing behaviors (Sec. \ref{sec:edit_velocity}). We then introduce Instruction-Level Similarity (ILS) to evaluate their predicted editing outcomes (Sec. \ref{sec:vlm_steer}) and dynamically select the velocity that best follows the instruction at each timestep (Sec. \ref{sec:velocity_control}). Finally, adaptive spatial masking restricts the selected update to regions that require modification, further preserving unedited content (Sec. \ref{sec:adaptive_mask}).

\subsection{Coupling-Controlled Editing Velocities}
\label{sec:edit_velocity}

FlowEdit uses a fixed source-anchored construction throughout the coupled editing stage, which strongly favors preservation but can suppress the requested semantic change. Rather than committing to a single coupling behavior, we expose a small set of editing velocities that relax source coupling in complementary ways. These velocities range from the original source-anchored update to locally decoupled, persistently decoupled, and extrapolated alternatives, providing different preservation--editing behaviors at each timestep. This velocity control set enables the editing trajectory to adapt its coupling behavior according to the current semantic requirement. We introduce the four editing velocities below.

\paragraph{Coupled Velocity.} The coupled velocity follows the original FlowEdit. As defined in Eq.~\eqref{eq:flowedit_xtar}, the target reference is anchored to the source reference: $X_t^{\mathrm{tar}} - X_t^{\mathrm{src}} = X_t^{\mathrm{edit}} - X^{\mathrm{src}}$. The source-target reference displacement is explicitly constrained by the current editing displacement, keeping the target reference anchored to the source. Under this source-anchored relation, the editing velocity is
\begin{equation*}
    V_t^{\mathrm{cpl}}
    = V_\theta(X_t^{\mathrm{tar}},t,\mathcal{C}^{\mathrm{tar}})
    - V_\theta(X_t^{\mathrm{src}},t,\mathcal{C}^{\mathrm{src}}),
\end{equation*}
which provides the most conservative candidate, favoring source preservation.

\paragraph{One-Step Decoupled Velocity.}
The coupled construction constrains the source-target reference displacement through Eq.~\eqref{eq:flowedit_xtar}. To locally relax this constraint, we propagate the coupled source and target references at $t + \Delta t$ independently by one ODE step under their respective conditions:
\begin{equation*}
\begin{aligned}
    \Tilde{X}_t^{\mathrm{src}} &= X_{t+\Delta t}^{\mathrm{src}} - V_\theta(X_{t+\Delta t}^{\mathrm{src}}, t + \Delta t, \mathcal{C}^{\mathrm{src}}) \Delta t, \\
    \Tilde{X}_t^{\mathrm{tar}} &= X_{t+\Delta t}^{\mathrm{tar}} - V_\theta(
    X_{t+\Delta t}^{\mathrm{tar}}, t + \Delta t, \mathcal{C}^{\mathrm{tar}}) \Delta t.
\end{aligned}
\end{equation*}
Since these propagated references are no longer reconstructed to satisfy the source-anchored relation at timestep $t$, their relative displacement can evolve according to the source- and target-conditioned flows. We then define the one-step decoupled velocity as
\begin{equation*}
    V_t^{\mathrm{dec}}
    = V_\theta(\Tilde{X}_t^{\mathrm{tar}}, t, \mathcal{C}^{\mathrm{tar}}) - V_\theta(\Tilde{X}_t^{\mathrm{src}}, t, \mathcal{C}^{\mathrm{src}}),
    \label{eq:decoupled_v}
\end{equation*}
which relaxes source coupling while remaining close to the coupled trajectory through a single-step propagation.

\paragraph{Extrapolated Velocity.} The coupled and one-step decoupled velocities represent different degrees of source coupling, with $V_t^{\mathrm{cpl}}$ favoring source preservation and $V_t^{\mathrm{dec}}$ allowing semantic deviation. To extend this preservation-editing spectrum, we extrapolate beyond the decoupled velocity along the direction induced by coupling relaxation:
\begin{equation*}
    V_t^{\mathrm{ext}} = 2V_t^{\mathrm{dec}} - V_t^{\mathrm{cpl}},
\end{equation*}
which provides a more aggressive candidate for edits that require stronger semantic modification while introducing no additional tunable parameter.

\paragraph{Persistent Decoupled Velocity.}
The one-step decoupled velocity relaxes the source-anchored relation locally, since the decoupled references are reconstructed from the coupled trajectory at each timestep. To capture the effect of sustained decoupling, we retain the decoupled source and target references across consecutive timesteps and propagate them recursively under their respective conditional flows. Let $\bar{X}_{t+\Delta t}^{\mathrm{src}}$ and $\bar{X}_{t+\Delta t}^{\mathrm{tar}}$ denote the persistent decoupled references from the previous step. We update them as
\begin{equation*}
\begin{aligned}
    \bar{X}_t^{\mathrm{src}} &= \bar{X}_{t+\Delta t}^{\mathrm{src}} - V_\theta(\bar{X}_{t+\Delta t}^{\mathrm{src}}, t + \Delta t, \mathcal{C}^{\mathrm{src}}) \Delta t, \\
    \bar{X}_t^{\mathrm{tar}} &= \bar{X}_{t+\Delta t}^{\mathrm{tar}} - V_\theta(\bar{X}_{t+\Delta t}^{\mathrm{tar}}, t + \Delta t, \mathcal{C}^{\mathrm{tar}}) \Delta t.
\end{aligned}
\end{equation*}
This recursive propagation does not reimpose the source-anchored relation, allowing the decoupling state established by one-step, extrapolated, or persistent decoupled updates to persist across timesteps. The corresponding velocity is
\begin{equation*}
    V_t^{\mathrm{pdec}} = V_\theta(\bar{X}_t^{\mathrm{tar}}, t, \mathcal{C}^{\mathrm{tar}}) - V_\theta(\bar{X}_t^{\mathrm{src}}, t, \mathcal{C}^{\mathrm{src}}),
\end{equation*}
which captures sustained coupling relaxation across consecutive timesteps, providing a temporally accumulated alternative to the local one-step decoupled velocity.

\paragraph{Velocity Control Set.}
At each timestep $t$, the control set $\mathcal{V}_t$ contains the coupled $V_t^{\mathrm{cpl}}$, one-step decoupled $V_t^{\mathrm{dec}}$, and extrapolated $V_t^{\mathrm{ext}}$ velocities. The persistent decoupled velocity $V_t^{\mathrm{pdec}}$ is additionally included whenever a decoupled reference trajectory is available from the previous timestep.

\subsection{Instruction-Level VLM Steering}
\label{sec:vlm_steer}

The velocity control set provides multiple editing behaviors at each timestep, but selecting among them requires a semantic signal that reflects the requested edit. Since human observers can naturally judge whether an image transformation follows an instruction, we use a vision-language model (VLM) as an automated proxy for such instruction-level judgment and define \emph{Instruction-Level Similarity} (ILS). ILS compares the intended transformation specified by the source image $X^{\mathrm{src}}$ and editing instruction $\mathcal{I}$ with the realized transformation observed from the source image $X^{\mathrm{src}}$ to the edited image $X^{\mathrm{edit}}$. Rather than generating textual descriptions and re-embedding them, we directly compare the corresponding VLM representations. 
Let $\phi^{(\ell)}(\cdot)$ denote the hidden representation at the first generation position of the VLM's $\ell$-th layer. We use the first generation position to reduce the influence of autoregressive decoding, where later token representations may increasingly reflect the generated sequence rather than the intended transformation. We select the final layer $L$, which empirically yields the strongest ILS performance, as verified in Sec.~\ref{sec:ablation_layer}. We define
\begin{equation}
    \mathrm{ILS}(X^{\mathrm{src}}, X^{\mathrm{edit}}, \mathcal{I})
    = \frac{\mathbf{h}_{\mathrm{int}}^\top \mathbf{h}_{\mathrm{edit}}}
    {\|\mathbf{h}_{\mathrm{int}}\|_2 \|\mathbf{h}_{\mathrm{edit}}\|_2},
\end{equation}
where
$\mathbf{h}_{\mathrm{int}} = \phi^{(L)}(X^{\mathrm{src}}, \mathcal{I})$
represents the VLM embedding of the intended transformation, and
$\mathbf{h}_{\mathrm{edit}} = \phi^{(L)}(X^{\mathrm{src}}, X^{\mathrm{edit}})$
represents that of the realized transformation.

A higher ILS indicates stronger agreement between the intended and realized transformations. We use this instruction-level signal to steer velocity selection throughout the editing trajectory.

\subsection{Dynamic Velocity Control}
\label{sec:velocity_control}

Given the velocity control set $\mathcal{V}_t$ and the instruction-level objective defined above, \method{} selects the velocity whose predicted editing outcome best matches the requested instruction. For each $V_t\in\mathcal{V}_t$, we form a clean-image preview via terminal ODE extrapolation:
\begin{equation}
    \hat{X}_{0|t}^\mathrm{edit}(V_t) = X_t^{\mathrm{edit}} - t V_t \approx X_t^{\mathrm{edit}} - \int_0^t V_\tau \mathrm{d} \tau,
    \label{eq:velocity_preview}
\end{equation}
and solve the optimization problem:
\begin{equation*}
    V_t^{*} = \argmax_{V_t\in\mathcal{V}_t}
    \mathrm{ILS}(X^{\mathrm{src}}, \hat{X}_{0|t}^\mathrm{edit}(V_t), \mathcal{I}).
\end{equation*}
After adaptive masking (Sec.~\ref{sec:adaptive_mask}),
the final velocity $V_t^{\mathrm{V2C}}$ is applied via one Euler step:
\begin{equation*}
    X_{t-\Delta t}^{\mathrm{edit}} = X_t^{\mathrm{edit}} - V_t^{\mathrm{V2C}} \Delta t.
\end{equation*}
Repeating this procedure at each timestep yields a closed-loop velocity controller that adapts the editing behavior throughout the trajectory.

\begin{table*}[t]
\caption{Quantitative comparison on MagicBrush and PIEbench. The best is marked in bold and the second best is underlined.}
\label{tab:main_table}
\centering
\setlength{\tabcolsep}{6pt}
\resizebox{\textwidth}{!}{%
\begin{tabular}{+l ^l ^c ^c ^c ^c ^c ^c ^c ^c ^c ^c}
    \toprule
    \multirow{2.5}{*}{\textbf{Method}} & \multirow{2.5}{*}{\textbf{Venue}} & \multicolumn{5}{c}{\textbf{MagicBrush}} & \multicolumn{5}{c}{\textbf{PIEbench}} \\
    \cmidrule(lr){3-7}\cmidrule(lr){8-12}
    ~ & ~ &
    \textbf{CLIP} $\uparrow$ &
    \textbf{$\text{IA}_\text{GPT}$} $\uparrow$ & \textbf{$\text{IA}_\text{Gemini}$} $\uparrow$ & \textbf{$\text{BP}_\text{PSNR}$} $\uparrow$ & \textbf{$\text{BP}_\text{SSIM}$} $\uparrow$ &
    \textbf{CLIP} $\uparrow$ &
    \textbf{$\text{IA}_\text{GPT}$} $\uparrow$ & \textbf{$\text{IA}_\text{Gemini}$} $\uparrow$ & \textbf{$\text{BP}_\text{PSNR}$} $\uparrow$ & \textbf{$\text{BP}_\text{SSIM}$} $\uparrow$ \\
    \midrule
    \multicolumn{12}{l}{\textbf{\textit{FLUX}}} \\
    RF-Edit~\cite{wang2025rfedit} & ICML'25 &0.2721	&1.48	&0.83	&\first{21.49}	&\first{0.6171} &0.2359	&1.83	&0.96	&\first{31.85}	&\first{0.9199} \\
    RF-Inversion~\cite{rout2025semantic} & ICLR'25 &0.2851	&4.12	&4.10	&17.16	&0.5301 &0.2562	&4.14	&4.12	&20.58	&0.7087 \\
    DNAEdit~\cite{xie2025dnaedit} & NeurIPS'25 &0.2825	&\second{5.67}	&4.61	&19.07	&0.5932 &0.2566	&\second{5.37}	&4.53	&25.31	&0.8831 \\
    FlowEdit~\cite{kulikov2025flowedit} & ICCV'25 &\second{0.2865}	&5.56	&\second{4.76}	&17.48	&0.5694 &0.2618	&5.32	&\second{4.78}	&21.82	&0.8411 \\
    FlowAlign~\cite{kim2026flowalign} & ICLR'26 &0.2778	&3.80	&3.27	&\second{20.25}	&\second{0.6162} &0.2485	&3.37	&2.81	&\second{27.82}	&\second{0.9184} \\
    \rowcolor{ourshl}
    \method{} (Ours) & WACV'27 &\first{0.2891}	& \first{6.07} & \first{5.36}	&17.44	& 0.5647 &\first{0.2636}	&\first{5.85} & \first{5.12} & 21.84 & 0.8414 \\
    \midrule
    \multicolumn{12}{l}{\textbf{\textit{Stable Diffusion 3}}} \\
    iRFDS~\cite{yang2025text_irfds} & ICLR'25 & 0.2648	& 3.88	& 2.47	& 16.77	&0.6401 &0.2556	&5.00	&3.55	& 19.57	& 0.7464 \\
    DNAEdit~\cite{xie2025dnaedit} & NeurIPS'25 & 0.2882	&5.74	&4.42	&\second{19.19}	&\second{0.5832} & 0.2644	&6.11	&5.08	&\second{25.67}	&\second{0.8847} \\
    FlowEdit~\cite{kulikov2025flowedit} & ICCV'25 &\first{0.2954}	& \second{6.28}	&\second{5.31}	&17.65	& 0.5585 &\first{0.2750}	&\second{6.63}	&\second{5.91}	& 22.17	&0.8397 \\
    FlowAlign~\cite{kim2026flowalign} & ICLR'26 & 0.2795	&5.10	& 4.30	&\first{30.53}	&\first{0.9374} & 0.2504	&5.13	&4.24	&\first{31.45}	&\first{0.9489} \\
    \rowcolor{ourshl}
    \method{} (Ours) & WACV'27 & \second{0.2948}	&\first{6.45}	&\first{5.33}	&17.64	& 0.5612 &\second{0.2746}	&\first{6.77}	&\first{6.00}	&22.49	&0.8491 \\
  \bottomrule
\end{tabular}
}
\end{table*}

\subsection{Adaptive Masking for Spatial Preservation}
\label{sec:adaptive_mask}

Since most editing instructions affect only part of the image, globally applying this update can unnecessarily alter regions that should remain unchanged. We therefore estimate the spatial extent of the predicted edit and use it to mask the velocity update. Given the terminal preview induced by the selected velocity $V_t^{*}$, we measure how much each pixel is expected to deviate from the source image:
\begin{equation*}
    E_t(p) = \| \hat{X}_{0|t}^{\mathrm{edit}} (V_t^{*})(p) - X^{\mathrm{src}}(p) \|_2,
\end{equation*}
where $p$ denotes a pixel location. Small edit magnitudes indicate regions that should remain unchanged, whereas larger values identify regions affected by the edit. We estimate an adaptive threshold $\eta_t$ from the smoothed edit-magnitude distribution by locating its steepest density drop, and define the edit mask as
\begin{equation*}
    M_t(p) = \mathbb{I} [E_t(p) > \eta_t].
\end{equation*}
We then apply $M_t$ to the selected velocity so that updates are concentrated on regions predicted to change while suppressing unnecessary modifications elsewhere, improving spatial preservation during editing.

\subsection{Overall Algorithm}
\label{sec:overall_algorithm}

To balance semantic editing and source preservation, we adopt a two-stage masking strategy guided by ILS. At each timestep, masking is considered only when the selected velocity $V_t^*$ achieves sufficient semantic alignment, and is applied only if it does not decrease the ILS score. This enables adaptive preservation without suppressing the requested edit. The complete algorithm is provided in the Appendix ~\ref{app:Algo.}.

\section{Experiments}

\begin{figure*}[t]
    \centering
    \includegraphics[width=\linewidth]{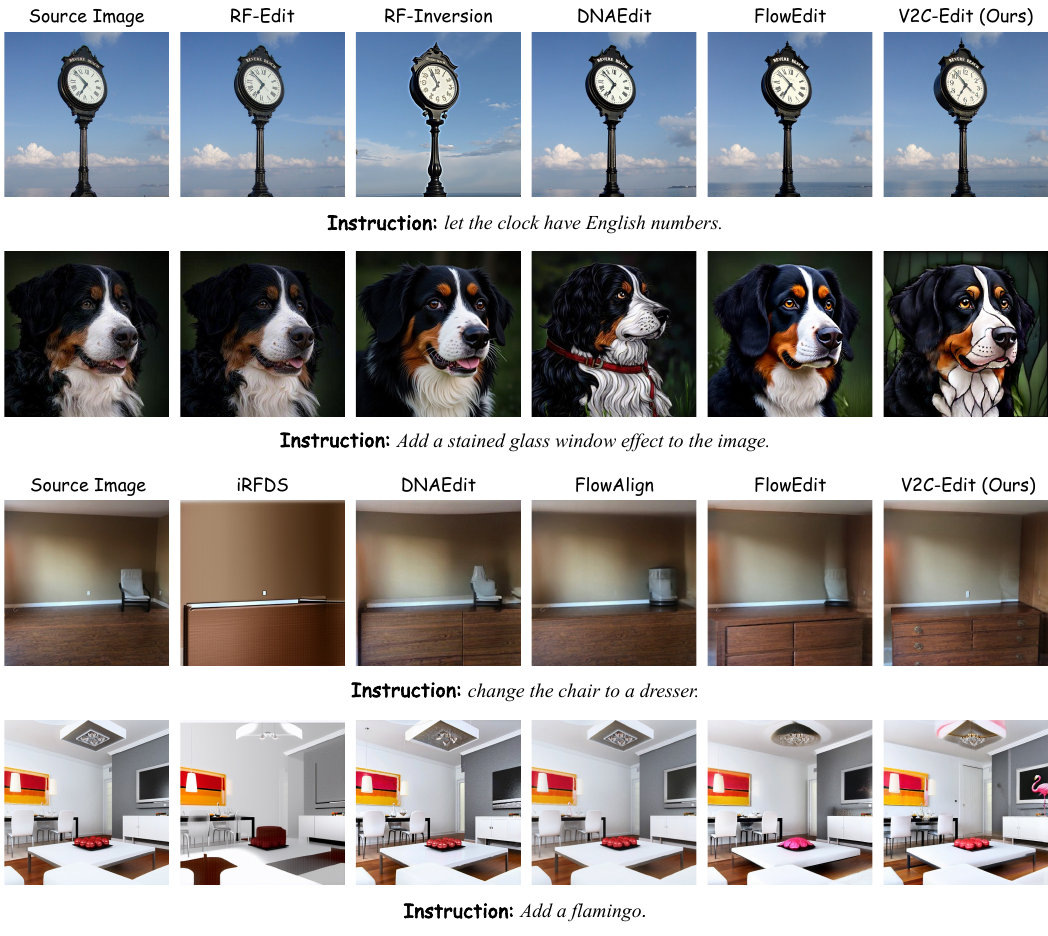}
    \caption{Qualitative comparisons of text-based image editing methods. Our method achieves the highest fidelity to the editing instruction while maintaining strong background consistency with the source image.}
    \label{fig:qual}
\end{figure*}

\subsection{Experimental Setup}

\paragraph{Datasets.}
We evaluate our method on MagicBrush~\cite{zhang2023magicbrush} and PIE-Bench~\cite{ju2024pnp}. MagicBrush is a large-scale, manually annotated dataset for instruction-guided editing of diverse real-world images. We adopt its single-turn test split of 535 samples and generate image captions with Qwen3-VL-8B~\cite{Qwen}. PIE-Bench consists of 700 synthetic and natural images covering 9 editing types, each paired with a source and target prompt.

\paragraph{Implementation and Compared Methods.}
We compare against representative training-free, flow-based editing methods, including RF-Edit~\cite{wang2025rfedit}, RF-Inversion~\cite{rout2025semantic}, DNAEdit~\cite{xie2025dnaedit}, FlowEdit~\cite{kulikov2025flowedit}, FlowAlign~\cite{kim2026flowalign}, and iRFDS~\cite{yang2025text_irfds}. As some implementations do not support both backbones, not every method is evaluated on both SD3-medium~\cite{sd3medium} and FLUX-dev~\cite{flux}. In particular, FlowAlign's official implementation only supports SD3-medium, so we additionally implement its FLUX-dev version ourselves. Given that FlowAlign and our method are both built upon FlowEdit, we adopt FlowEdit's~\cite{kulikov2025flowedit} hyperparameter setup for these two methods to ensure a fair comparison; the remaining baselines are run with their official implementations and default settings.

\subsection{Evaluation Protocol}
\label{sec:evaluation_metric}

\paragraph{Semantic Alignment.}
Our method leverages VLM hidden states as guidance during editing; using ILS as the sole semantic evaluation metric could therefore introduce evaluation bias. To mitigate this, we introduce GPT-4.1~\cite{openaigpt41} and Gemini 3.6 Flash~\cite{gemini36} as independent human-proxy evaluators, scoring results along three dimensions: \textbf{Instruction Alignment (IA)}, which measures whether the edited result satisfies the editing instruction; \textbf{Subject Preservation (SP)}, which measures how well the edited subject is preserved; and \textbf{Background Preservation (BP)}, which evaluates background consistency through the VLM's holistic judgment, complementing the pixel-level background preservation metrics introduced below. Each dimension is scored on a scale of 0 to 10, with the scoring prompts provided in Appendix ~\ref{app:prompt_tmp}. As shown in Fig.~\ref{fig:clip_failure}, the CLIP score, computed as the cosine similarity between the CLIP embedding of the edited image and that of the target caption, fails to faithfully capture editing quality. We nonetheless report the CLIP score alongside the VLM-based IA score to provide two perspectives on semantic alignment.
\paragraph{Background Preservation.}
To quantify background preservation, we compute PSNR and SSIM restricted to the unedited region, based on the editing mask provided by MagicBrush and PIE-Bench. PSNR measures pixel-level fidelity, while SSIM measures structural similarity between source and edited images; these two metrics jointly reflect how well the unedited regions are preserved.

\subsection{Results}
As shown in Tab.~\ref{tab:main_table}, our method achieves the best IA while maintaining preservation comparable to FlowEdit, indicating that our instruction-alignment gains do not come at the expense of background preservation. RF-Edit and FlowAlign attain the strongest preservation scores but substantially lower IA, exposing a trade-off that our method mitigates.
Fig.~\ref{fig:qual} shows qualitative results on FLUX and SD3. In the first case for each model, baselines fail to achieve the intended semantic edit, while our method satisfies the instruction with preservation comparable to other approaches. In the second case, baselines introduce visible changes but miss the target semantics, whereas our method achieves the most faithful stained-glass stylization and correctly incorporates the flamingo, while other methods produce inaccurate or misdirected edits. These results demonstrate that our method combines strong instruction-following with competitive background preservation across diverse editing scenarios.

\begin{table}[t]
\centering
\caption{User study results. Percentage indicates the proportion of participant selections received by each method.}
\label{tab:user_study}
\resizebox{\linewidth}{!}{%
\begin{tabular}{lcccc}
  \toprule
  \textbf{Method} & \textbf{DNAEdit}~\cite{xie2025dnaedit} & \textbf{FlowEdit}~\cite{kulikov2025flowedit} & \textbf{FlowAlign}~\cite{kim2026flowalign} & \textbf{\method{} (Ours)} \\
  \midrule
  Percentage & 11.0\% & 16.7\% & 18.5\% & \first{53.8\%} \\
  \bottomrule
\end{tabular}
}
\end{table}

\subsection{User Study}
\label{sec:user_study}
Since image editing is inherently a one-to-many task without a unique ground truth, quantitative metrics alone may not fully capture perceptual editing quality. To complement our quantitative evaluation, we conduct a user study on 20 randomly sampled editing instances from MagicBrush~\cite{zhang2023magicbrush}, comparing DNAEdit~\cite{xie2025dnaedit}, FlowEdit~\cite{kulikov2025flowedit}, FlowAlign~\cite{kim2026flowalign}, and our \method{}. For each instance, participants are presented with four edited results in randomized order and asked to select the one that best follows the editing instruction while preserving regions that should remain unchanged. As shown in Tab.~\ref{tab:user_study}, our \method{} achieves the highest user preference, demonstrating its effectiveness in balancing editing fidelity and source preservation. Further details are provided in the Appendix ~\ref{app:user_study_supple}.

\subsection{Ablation Study}
All ablation studies are conducted on FLUX~\cite{flux} for computational efficiency, as it requires fewer inference steps ($T=28$) than SD3~\cite{sd3medium} ($T=50$).

\paragraph{Editing Velocity Candidates.}
We investigate the contribution of different editing velocity candidates by progressively introducing Decoupled, Extrapolated, and Persistent Decoupled directions to the Coupled baseline. As shown in Tab.~\ref{tab:ablation_candidate}, incorporating additional candidates progressively improves instruction alignment (IA), from 5.69 to 6.07, at the cost of reduced source preservation. These results highlight the preservation–editing trade-off and demonstrate the importance of diverse editing directions in enabling stronger semantic modifications.

\paragraph{VLM Hidden State Selection.}
\label{sec:ablation_layer}

We evaluate ILS representations using Qwen3-VL-8B-Instruct~\cite{Qwen} on the MagicBrush test set, comparing source images with their human-edited ground truths. As shown in Tab.~\ref{tab:ablation_vlm_layer}, Anomaly rate measures the percentage of cases where the source scores at least as high as the ground truth, while mean score gap measures the average score difference (ground truth minus source). Target-caption CLIP similarity exhibits a 35.88\% anomaly rate and a small mean score gap (0.0122), indicating limited discrimination between edited and unedited images. First-layer VLM representations yield a 100\% anomaly rate, whereas final-layer representations reduce it to 16.82\% and 2.43\% for the final and first tokens, respectively. The final-layer, first-token representation achieves the lowest anomaly rate and largest mean score gap, indicating superior discrimination between edited and unedited images. We therefore adopt this configuration for ILS.

\begin{table}[t]
\caption{Ablation study on editing direction candidates.}
\label{tab:ablation_candidate}
\centering
\resizebox{\linewidth}{!}{
\begin{tabular}{ccccccccc}
    \toprule
    $\mathbf{V}_t^\mathrm{cpl}$ & $\mathbf{V}_t^\mathrm{dec}$ & $\mathbf{V}_t^\mathrm{ext}$ & $\mathbf{V}_t^\mathrm{pdec}$ & \textbf{CLIP} $\uparrow$ & \textbf{$\text{IA}_\text{GPT}$} $\uparrow$ & \textbf{$\text{BP}_\text{PSNR}$} $\uparrow$ & \textbf{$\text{BP}_\text{SSIM}$} $\uparrow$ \\
    \midrule
    \checkmark & \xmark & \xmark & \xmark & 0.2864 & 5.69 & \first{17.78} & \first{0.5787} \\ 
    \checkmark & \checkmark & \xmark & \xmark & \second{0.2888} & \second{5.91} & \first{17.78}	& \second{0.5750} \\
    \checkmark & \checkmark & \checkmark & \xmark & 0.2879 & \second{5.91} & \second{17.57}	& 0.5694 \\
    \checkmark & \checkmark & \checkmark  & \checkmark & \first{0.2891} & \first{6.07} &  17.44	& 0.5647 \\
  \bottomrule
\end{tabular}
}
\end{table}

\begin{table}[t]
    \caption{Comparison of metrics across token positions and VLM layers on the MagicBrush. Anomaly rate measures the percentage of cases where the source scores at least as high as the ground truth. Mean score gap measures the average score difference.}
    \label{tab:ablation_vlm_layer}
    \centering
    \resizebox{\linewidth}{!}{%
    \begin{tabular}{ccc|cc}
    \toprule
    \textbf{Metric} & \textbf{Token} & \textbf{Layer} & \textbf{Anomaly Rate} $\downarrow$ & \textbf{Mean Score Gap} $\uparrow$ \\
    \midrule
    CLIP & - & - & 35.88\% & 0.0122 \\
    \midrule
    \multirow{4}{*}{ILS} & First & First & 100\% & 0 \\
    ~ & Final  & First & 100\% & 0 \\    
    ~ & Final  & Final & 16.82\% & 0.0533 \\     
    ~ & First & Final & \textbf{2.43\%} & \textbf{0.2044} \\
    \bottomrule
    \end{tabular}
    }
\end{table}

\section{Conclusion}

We presented \method{}, a training-free method for mitigating the \emph{zero-edit} failure of FlowEdit. Our analysis shows that source coupling governs the preservation--editing trade-off, motivating per-timestep velocity control instead of a fixed coupling schedule. \method{} dynamically selects among complementary editing velocities using VLM-based Instruction-Level Similarity, while adaptive masking preserves unedited regions. Experiments on MagicBrush and PIE-Bench show improved instruction alignment over FlowEdit with competitive structural preservation. Our results show that dynamic, instruction-aware velocity control provides an effective way to balance semantic editing and source preservation in inversion-free image editing.

\section*{Acknowledgements}

This work was supported by the National Science and Technology Council (NSTC), Taiwan, under Grants NSTC 115-2634-F-001-006 and NSTC 114-2221-E-002-182-MY3. We thank the National Center for High-performance Computing (NCHC) of the National Applied Research Laboratories (NARLabs), Taiwan, for providing computational and storage resources. We acknowledge the generous support of the National Science and Technology Center for Disaster Reduction (NCDR) and the Industrial Technology Research Institute (ITRI), Taiwan.

{
    \small
    \bibliographystyle{ieeenat_fullname}
    \bibliography{main}
}

\clearpage
\appendix
\renewcommand\thesection{\Alph{section}}
\setcounter{section}{0}

\twocolumn[
    \vspace{1cm}
    \begin{center}
        \Large{\textbf{[Appendix] VLM as an Editing Pilot: VLM-Steered         Velocity Control for Flow-Based Image Editing}}
    \end{center}
    \vspace{1cm}
]

\DoToC

\section{Algorithm Details}
\label{app:Algo.}

We provide the complete pseudocode for \method{}. Algo.~\ref{alg:v2c-edit_simple} summarizes the overall editing process, Algo.~\ref{alg:candidate} constructs the coupling-controlled velocity candidates, and Algo.~\ref{alg:masking} performs two-stage adaptive masking. We define
\begin{align*}
\mathcal{P}_{t\to t'}(X, T) &= X - (t - t')\, V_\theta(X, t, T) \\
\mathcal{D}_t(X, Y) &= V_\theta(X, t, T_{\mathrm{tar}}) - V_\theta(Y, t, T_{\mathrm{src}})\\
\hat{X}_{0|t}^{\mathrm{edit}}(V) &= X_t^{\mathrm{edit}} - t V
\end{align*}
denoting a single Euler propagation step, the source-target velocity difference, and the terminal-preview operator, respectively.
The subroutines share model conditions and intermediate states across timesteps. Persistent references are updated according to the selected velocity type before masking and reset when the coupled velocity is selected. In addition, the histogram used for adaptive masking is smoothed before locating the steepest density drop.

\begin{algorithm}[t]
\caption{V2C-Edit}
\label{alg:v2c-edit_simple}
\begin{algorithmic}[1]
\Require source image $X^{\mathrm{src}}$; source / target / instruction text conditions $T_{\mathrm{src}}, T_{\mathrm{tar}}, T_{\mathrm{instr}}$; pretrained flow model $V_\theta$; decreasing timestep schedule $\{t_i\}_{i=0}^{n_{\max}}$; semantic threshold $\tau$
\Ensure $X_{t_0}^{\mathrm{edit}}$

\State $X_{t_{n_{\max}}}^{\mathrm{edit}} \leftarrow X^{\mathrm{src}}$
\State $\bar X_{t_{n_{\max}}}^{\mathrm{src}} \leftarrow \varnothing$

\State $\bar X_{t_{n_{\max}}}^{\mathrm{tar}} \leftarrow \varnothing$

\For{$i=n_{\max}$ \textbf{to} $1$}
    \State $(\mathcal V_{t_i},\tilde X_{t_i}^{\mathrm{src}},\tilde X_{t_i}^{\mathrm{tar}})
    \leftarrow \Call{ConstructCandidate}{t_i}$
    
    \State $\displaystyle (V_{t}^*,s_{t_i}) \leftarrow
    \argmax_{V\in\mathcal V_{t_i}}
    \mathrm{ILS}(X^{\mathrm{src}},
    \hat X_{0|t_i}^{\mathrm{edit}}(V),T_{\mathrm{instr}})$
    
    \State $k \leftarrow \operatorname{type}(V_{t}^*)$
    \State $V_{\mathrm{V2C}}\leftarrow
    \Call{AdaptiveMasking}{V_{t}^*,s_{t_i},t_i,\tau}$
    
    \State $X_{t_{i-1}}^{\mathrm{edit}}\leftarrow
    X_{t_i}^{\mathrm{edit}}+
    (t_{i-1}-t_i)V_{\mathrm{V2C}}$

    \If{$k\in\{\mathrm{dec},\mathrm{ext}\}$}
        \State $\bar X_{t_{i-1}}^{\mathrm{src}} \leftarrow \mathcal{P}_{t_i \to t_{i-1}}(\tilde X_{t_i}^{\mathrm{src}}, T_{\mathrm{src}})$
        \State $\bar X_{t_{i-1}}^{\mathrm{tar}} \leftarrow \mathcal{P}_{t_i \to t_{i-1}}(\tilde X_{t_i}^{\mathrm{tar}}, T_{\mathrm{tar}})$
    \ElsIf{$k=\mathrm{pdec}$}
        \State $\bar X_{t_{i-1}}^{\mathrm{src}} \leftarrow \mathcal{P}_{t_i \to t_{i-1}}(\bar X_{t_i}^{\mathrm{src}}, T_{\mathrm{src}})$
        \State $\bar X_{t_{i-1}}^{\mathrm{tar}} \leftarrow \mathcal{P}_{t_i \to t_{i-1}}(\bar X_{t_i}^{\mathrm{tar}}, T_{\mathrm{tar}})$
    \Else
        \State $\bar X_{t_i-1}^{\mathrm{src}}
        \leftarrow \varnothing$
        \State $\bar X_{t_i-1}^{\mathrm{tar}}
        \leftarrow \varnothing$
    \EndIf
\EndFor
\State \Return $X_{t_0}^{\mathrm{edit}}$
\end{algorithmic}
\end{algorithm}

\section{Implementation Details}

\paragraph{VLM Prompt Design for Instruction-Level Similarity.}
Instruction-level similarity (ILS) extracts hidden representations from two separate VLM queries, each formulated as a 20-word instruction to ensure consistent comparison between intended and realized transformations. The complete query templates are provided at the end of the Appendix.

\paragraph{Human-Proxy Evaluation Prompt}
We use the same evaluation prompt for both GPT-4.1~\cite{openaigpt41} and Gemini 3.6 Flash~\cite{gemini36}, as described in Sec.~6.2. The complete prompt template is provided below.

\begin{figure*}[t]
    \centering
    \includegraphics[width=0.96\linewidth]{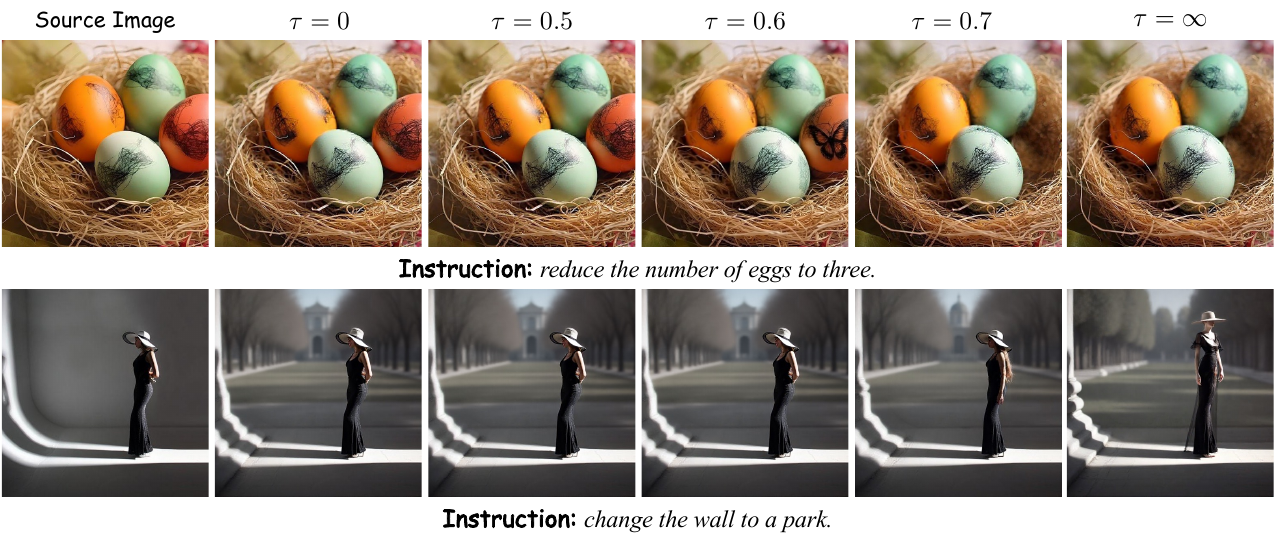}
    \caption{
    \textbf{Qualitative effect of the semantic threshold $\tau$.}
    \textit{Top:} Lower thresholds ($\tau \leq 0.6$) trigger premature masking, leading to incomplete edits, whereas the default $\tau=0.7$ successfully reduces the number of eggs to three. \textit{Bottom:} Although all settings satisfy the instruction, disabling masking ($\tau=\infty$) alters the person's appearance and clothing, while $\tau=0.7$ better preserves the source content.
    }
    \label{fig:ablation_mask_qualitative}
\end{figure*}

\begin{table}[t]
\caption{\textbf{Ablation on the semantic threshold.} $\tau=0.0$ applies masking at every timestep, $\tau=\inf$ disables masking entirely, and intermediate values gate masking based on the semantic score.}
\label{tab:ablation_masking}
\centering
\setlength{\tabcolsep}{4pt}
\resizebox{\linewidth}{!}{%
\begin{tabular}{c|cccc}
    \toprule
    \textbf{Confidence} $\tau$ &\textbf{CLIP} $\uparrow$ & \textbf{$\text{IA}_\text{GPT}$} & \textbf{$\text{BP}_\text{PSNR}$} $\uparrow$ & \textbf{$\text{BP}_\text{SSIM}$} $\uparrow$ \\
    \midrule
    0.0 & 0.2877 & 5.72 & \first{18.36}&\first{0.5864} \\ 
    0.4 & 0.2877 & 5.76 & \first{18.36} & \first{0.5864}\\
    0.5 & 0.2876 & 5.74 & \second{18.34} & \second{0.5860}\\
    0.6 & 0.2881 & \second{5.96} & 17.91 & 0.5762 \\
    0.7 &\first{0.2891}	&\first{6.07}	&17.44 & 0.5647 \\
    $\inf$ &\second{0.2886}	&\first{6.07} & 17.08	 &0.5535 \\
  \bottomrule
\end{tabular}
}
\end{table}

\begin{figure*}[t]
  \centering
  \includegraphics[width=0.96\linewidth]{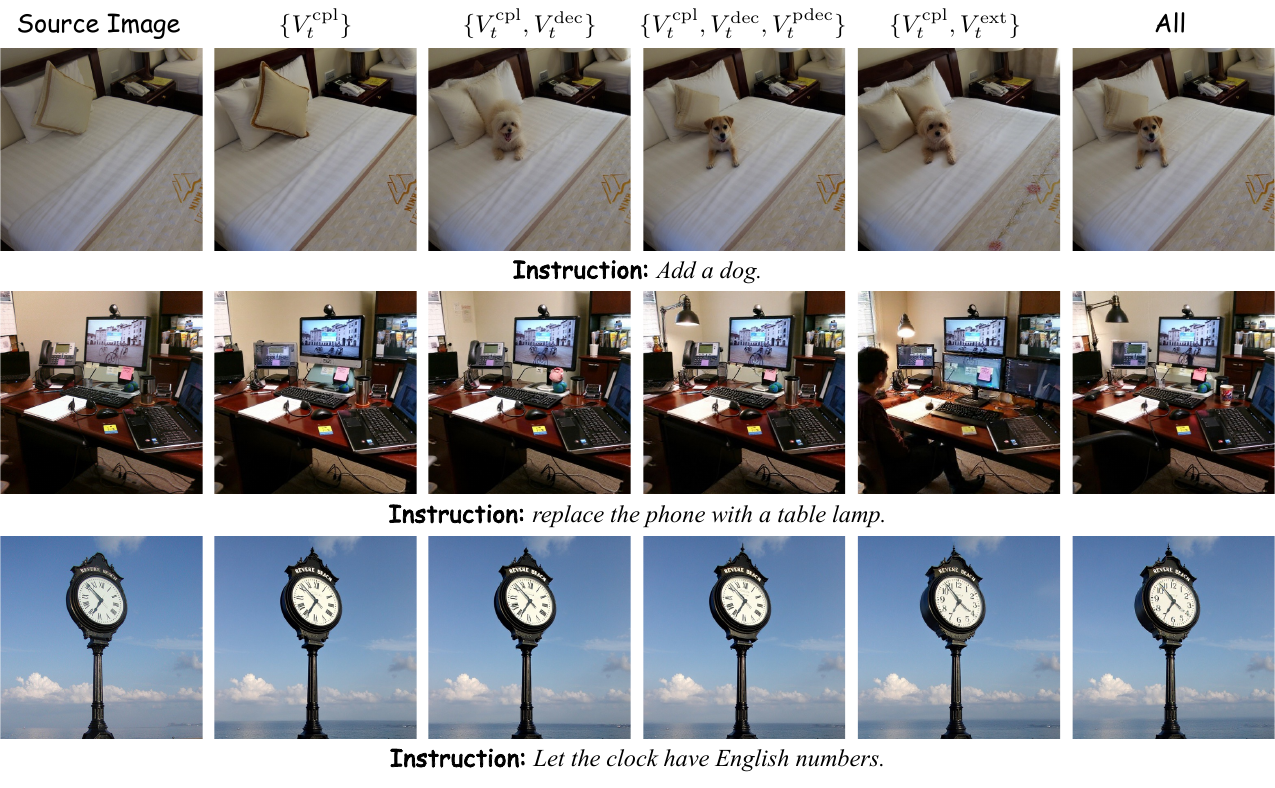}
        \caption{\textbf{Qualitative effect of progressively adding velocity candidates.}
        Top: the coupled velocity produces a zero-edit result, while introducing a relaxed velocity enables the dog to be added. Middle: local one-step relaxation remains insufficient to replace the phone with a table lamp, whereas persistent decoupling successfully completes the edit; extrapolation also achieves the requested change but introduces larger unintended modifications. Bottom: extrapolation is required to correctly convert the clock numerals to English digits. Combining all candidates allows the controller to select the appropriate degree of coupling relaxation for different editing requirements.}
  \label{fig:ablation_candidate}
\end{figure*}

\begin{figure*}[t]
  \centering
  \includegraphics[width=\linewidth]{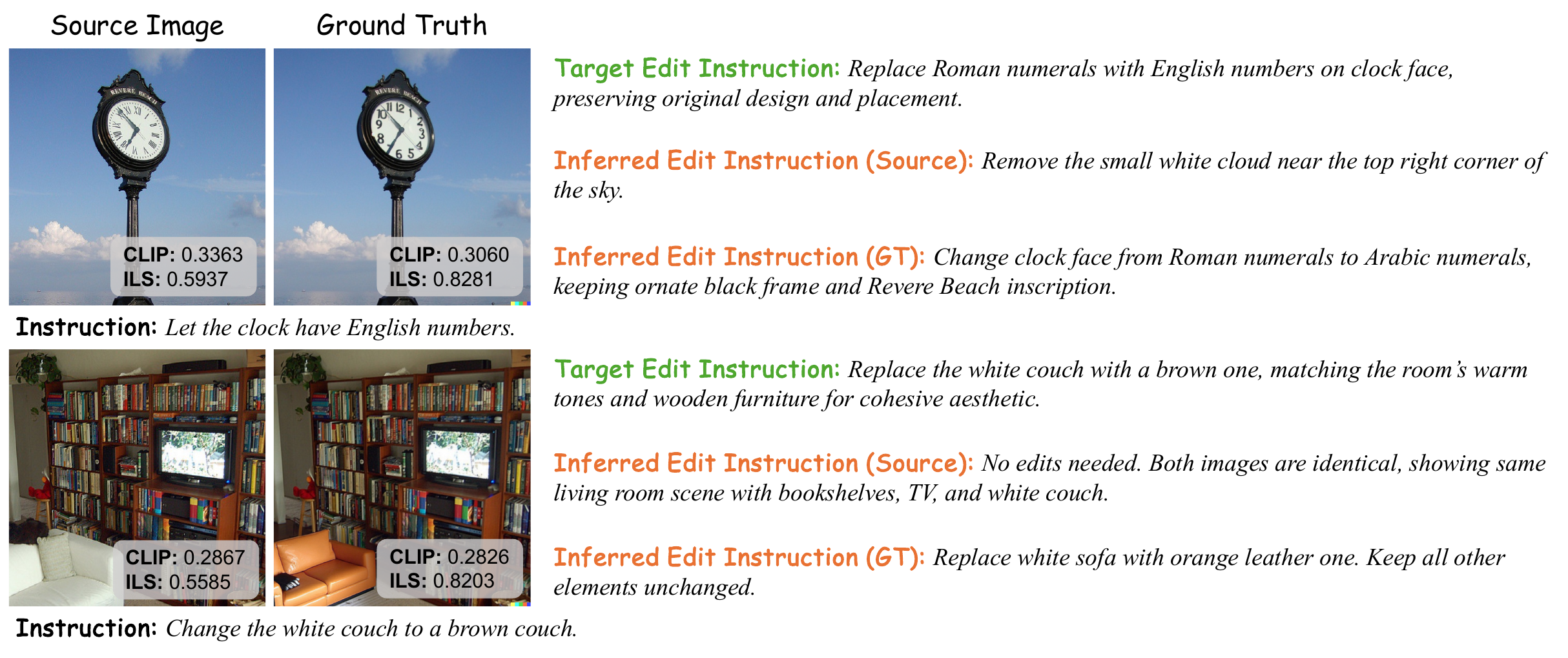}
          \caption{\textbf{Qualitative examples of ILS on MagicBrush source-ground truth pairs.} For each example, the VLM generates a target edit instruction from the source image and instruction, and an inferred edit instruction from either the source image itself (source-to-source, no actual edit) or the ground truth image (source-to-GT, the intended edit).}
  \label{fig:ils_qualitative}
\end{figure*}

\section{More Analysis}

\subsection{Masking Threshold}
\label{app:masking_threshold_qual}

As shown in Tab.~\ref{tab:ablation_masking} and Fig.~\ref{fig:ablation_mask_qualitative}, the semantic threshold $\tau$ governs the trade-off between instruction alignment and source preservation. Lower thresholds trigger earlier masking, improving preservation but potentially limiting semantic edits. In Fig.~\ref{fig:ablation_mask_qualitative} (top), $\tau \leq 0.6$ results in incomplete edits, whereas $\tau=0.7$ correctly reduces the number of eggs to three. Conversely, disabling masking ($\tau=\infty$) permits stronger edits but introduces unintended changes to the source content, as shown in the bottom example. We therefore adopt $\tau=0.7$ as the default, achieving a favorable balance between instruction alignment and source preservation.

\begin{figure*}[t]
    \centering
    \begin{subfigure}{0.48\linewidth}
        \centering
        \includegraphics[width=\linewidth]{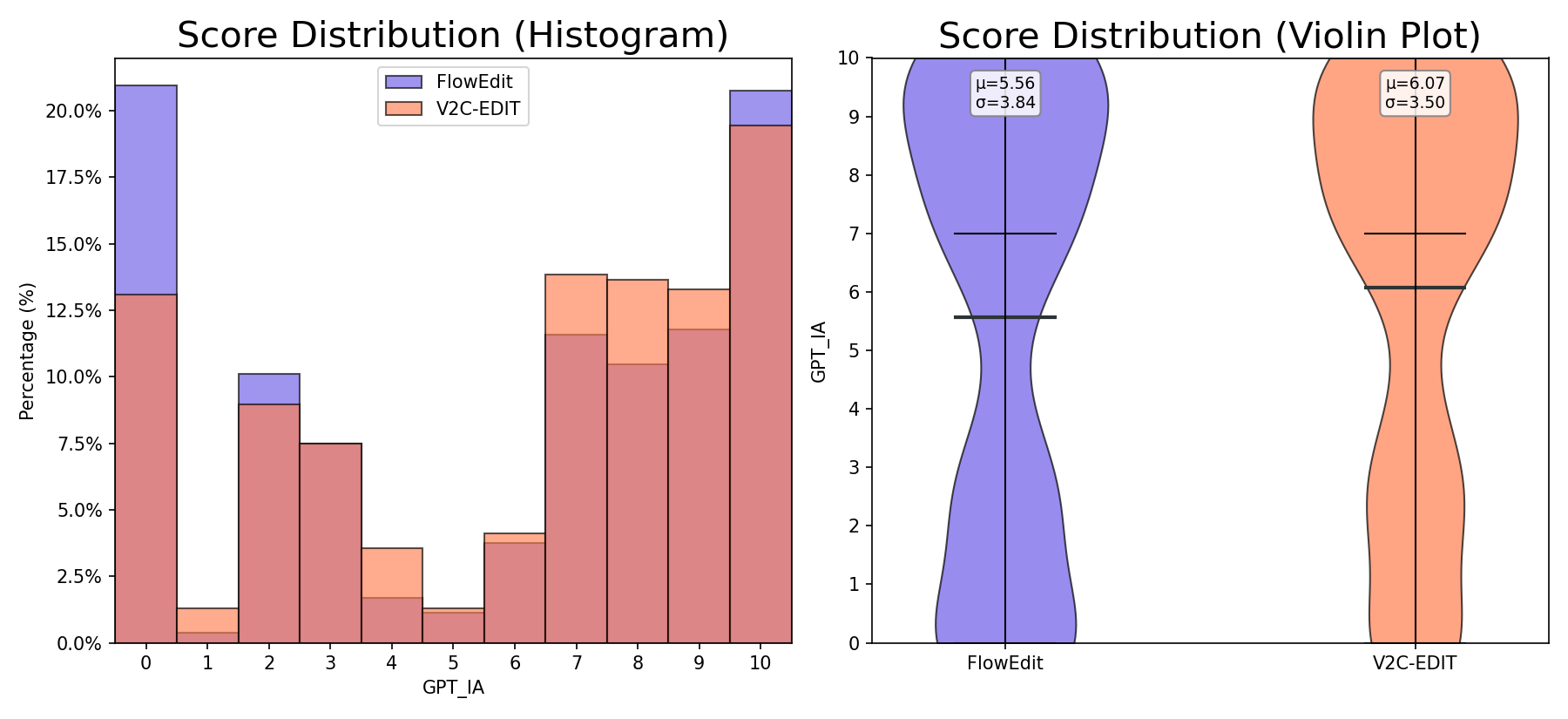}
        \caption{MagicBrush}
        \label{fig:ia_violin_magic_flux}
    \end{subfigure}
    \hfill
    \begin{subfigure}{0.48\linewidth}
        \centering
        \includegraphics[width=\linewidth]{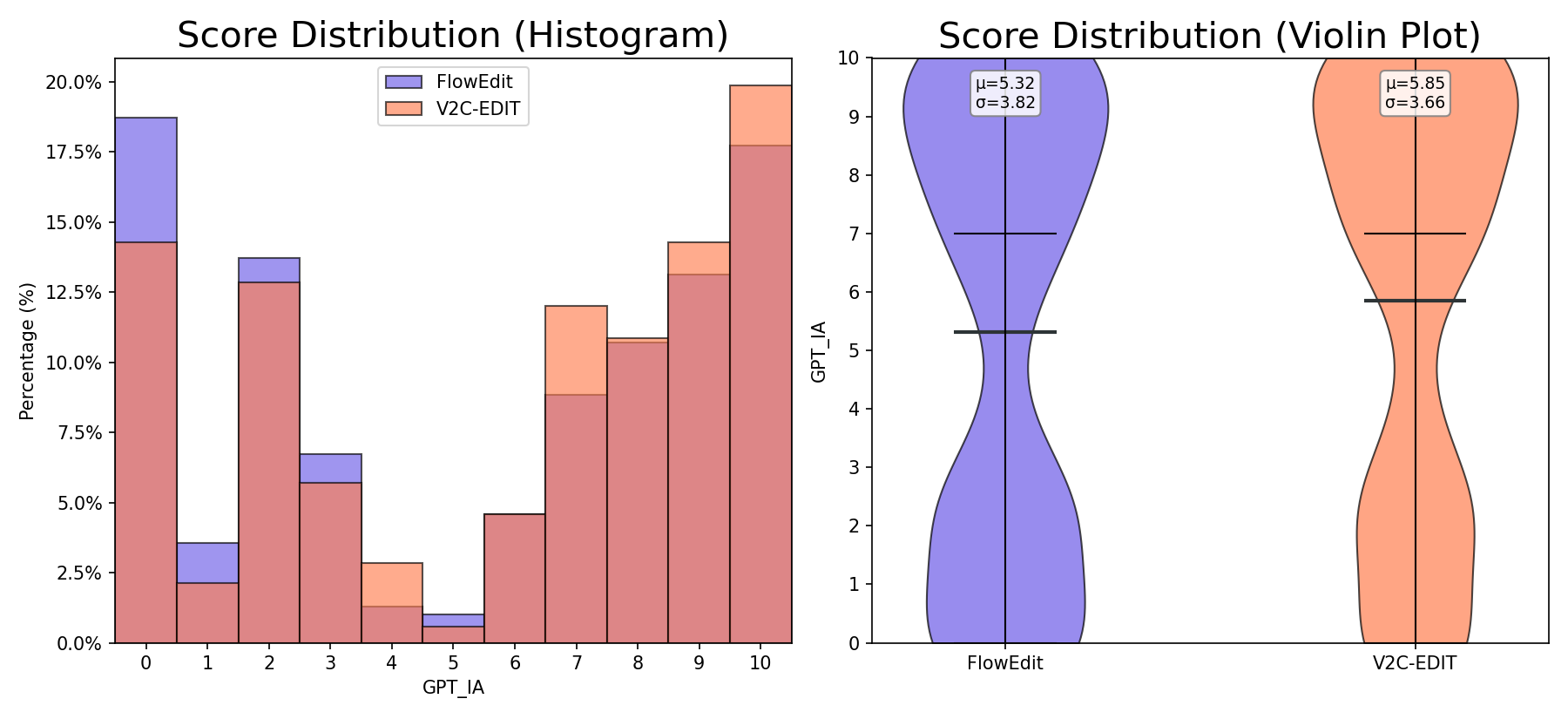}
        \caption{PIE-Bench}
        \label{fig:ia_violin_pie_flux}
    \end{subfigure}
    \caption{\textbf{Distribution of IA$_{\mathrm{GPT}}$ scores on FLUX.}
    Per-sample IA$_{\mathrm{GPT}}$ distributions of FlowEdit and \method{} on MagicBrush and PIE-Bench are shown using histograms and violin plots. On both benchmarks, \method{} substantially reduces the concentration of samples in the low-score region and redistributes more samples toward medium-to-high IA$_{\mathrm{GPT}}$ scores. On MagicBrush, this shift is mainly reflected in fewer low-alignment cases and increased density across the intermediate and high-score ranges, although the number of samples receiving the maximum score of 10 is slightly lower than that of FlowEdit. On PIE-Bench, \method{} similarly reduces low-score outcomes while showing a broader concentration in the upper-score region. The violin plots reflect the same redistribution, with less density near the lower end and greater concentration at higher IA$_{\mathrm{GPT}}$ values.}
    \label{fig:ia_violin}
\end{figure*}

\subsection{Coupling-Controlled Editing Velocity}
\label{app:candidate_ablation_qual}
To complement the quantitative ablation in Tab.~\ref{tab:ablation_candidate} of the main paper, Fig.~\ref{fig:ablation_candidate} illustrates how different coupling-controlled velocities provide complementary editing behaviors. The coupled velocity is the most conservative and can fail when stronger semantic deviation is required. One-step decoupling improves editability through local coupling relaxation, while persistent decoupling is more effective for edits that require sustained deviation across timesteps. Extrapolation provides the strongest semantic modification, but can also introduce unnecessary changes when such aggressive editing is not needed. By jointly considering all candidates, our method can adapt the degree of coupling relaxation to each editing trajectory, improving instruction alignment without consistently incurring the preservation cost of more aggressive velocities.

\subsection{Instruction-Level Similarity}
\label{app:ils_qual}
We provide qualitative examples supporting the layer selection analysis in Sec. 6.5, shown in Fig.~\ref{fig:ils_qualitative}. When the inferred instruction is derived from the source image itself, it reports minimal or no change, yielding a low ILS score; when derived from the ground truth image, it closely matches the target instruction in content, yielding a high ILS score. This confirms that ILS correctly reflects semantic agreement between the intended and realized transformations.

\section{More Results}
\label{app:quant_results}

To further analyze instruction alignment, Fig.~\ref{fig:ia_violin} compares the IA$_{\mathrm{GPT}}$ score distributions of FlowEdit~\cite{kulikov2025flowedit} and our method on FLUX. Across MagicBrush~\cite{zhang2023magicbrush} and PIE-Bench~\cite{ju2024pnp}, our method shifts the distributions toward higher scores, primarily by reducing severe low-alignment failures rather than increasing near-perfect edits. Combined with the preservation results in Tab.~\ref{tab:main_table} of the main paper, these findings demonstrate that our method improves instruction alignment while maintaining source preservation, achieving a more favorable preservation–editing trade-off.

\begin{table}[t]
\caption{Comparison with brute-force $t_\mathrm{cs}$ search on FLUX. For each test sample, all possible values of $t_\mathrm{cs}$ are exhaustively evaluated, and the result with the highest ILS score is selected, representing the best achievable outcome of FlowEdit's original mechanism when paired with ILS.}
\label{tab:ablation_brute_force}
\centering
\setlength{\tabcolsep}{4pt}
\resizebox{\linewidth}{!}{%
\begin{tabular}{lccccc}
    \toprule
    \textbf{Method} & \textbf{CLIP} $\uparrow$ & \textbf{$\text{IA}_\text{GPT}$} $\uparrow$ & \textbf{$\text{BP}_\text{PSNR}$} $\uparrow$ & \textbf{$\text{BP}_\text{SSIM}$} $\uparrow$ \\
    \midrule
    FlowEdit ($t_\mathrm{cs}=0$) &0.2865	&5.56 &\first{17.48} &\first{0.5694}\\
    FlowEdit (Brute force $t_\mathrm{cs}$) &\first{0.2892} &\first{6.22} & 16.18	& 0.5190 \\
    \rowcolor{ourshl}
    \method{} (Ours) &\second{0.2891}	&\second{6.07}	&\second{17.44}	&\second{0.5647} \\
  \bottomrule
\end{tabular}
}
\end{table}

\section{Computational Cost}
\label{app:brute_force}

\begin{figure}
    \centering
    \includegraphics[width=\linewidth]{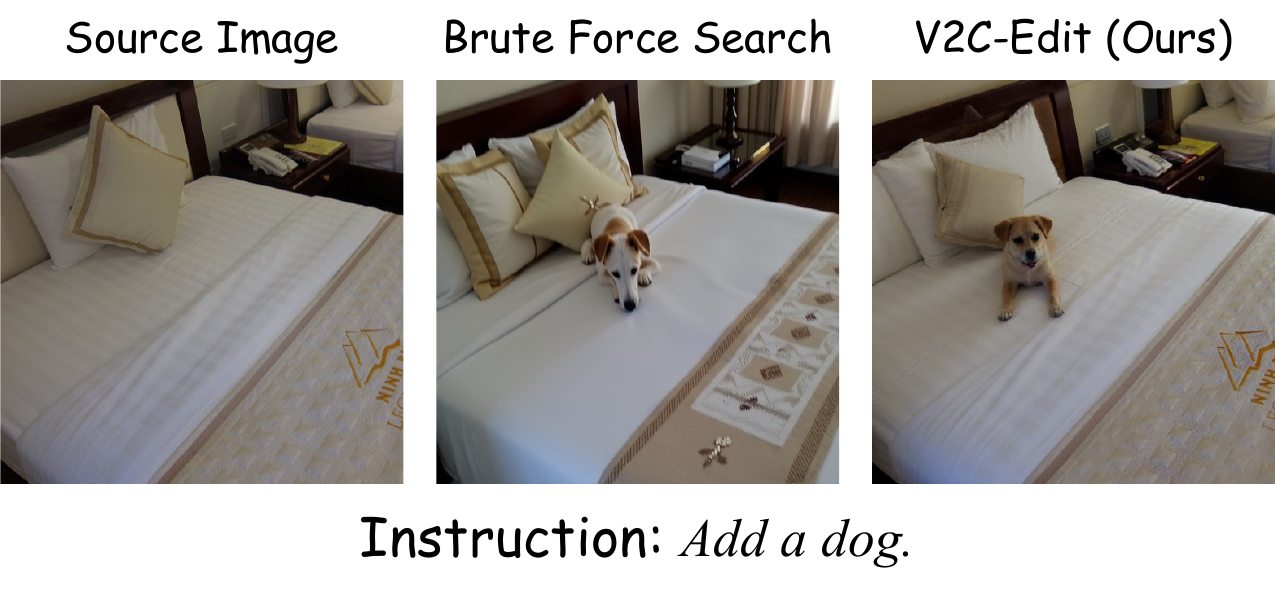}
    \caption{\textbf{Qualitative comparison with Brute-Force $t_{\mathrm{cs}}$ search.} From left to right: source image, Brute Force, and Ours. Both methods satisfy the instruction, but Brute Force introduces noticeably more changes to the background, while our method preserves the unedited regions much more faithfully.}
    \label{fig:brute_force_qualitative}
\end{figure}

To assess whether exhaustive optimization of a fixed coupling switch point can replace per-timestep adaptive control, we search all possible $t_{\mathrm{cs}}$ values and select the result with the highest ILS score for each sample. As shown in Tab.~\ref{tab:ablation_brute_force}, brute force achieves slightly higher instruction alignment (IA) but substantially worse background preservation. Fig.~\ref{fig:brute_force_qualitative} further illustrates this trade-off, where brute force introduces noticeable background changes despite successfully adding the requested dog. As reported in Tab.~\ref{tab:ablation_brute_force_time}, Brute Force requires 79.99s per image, compared with 26.44s for ours, with identical peak VRAM usage (53 GB). Our method thus achieves comparable instruction alignment with better background preservation at approximately one-third of the inference time, demonstrating the efficiency of adaptive per-timestep velocity control.

\begin{table}[t]
\centering
\caption{Average inference time and peak VRAM usage per image,
measured on an NVIDIA H100 80GB GPU over 50 test samples.}
\label{tab:ablation_brute_force_time}
\resizebox{\linewidth}{!}{%
\begin{tabular}{lcc}
    \toprule
    \textbf{Method} & \textbf{Time (s) $\downarrow$} & \textbf{VRAM (GB) $\downarrow$} \\
    \midrule
    FlowEdit ($t_\mathrm{cs}=0$) & 4.00  & 36 \\
    FlowEdit (Brute force $t_\mathrm{cs}$) & 79.99 & 53 \\
    \rowcolor{ourshl}
    \method{} (Ours) & 26.44 & 53 \\
  \bottomrule
\end{tabular}
}
\end{table}

\section{Prompt Template}
\label{app:prompt_tmp}
We provide the prompt templates used for ILS computation and human-proxy evaluation. For ILS, two VLM queries capture the intended and realized transformations. The former takes the source image and editing instruction, while the latter takes the source and edited images. Both request a 20-word description, but ILS computes cosine similarity between VLM hidden representations rather than generated text. For human-proxy evaluation, GPT-4.1 and Gemini 3.6 Flash independently assess Instruction Alignment (IA), Subject Preservation (SP), and Background Preservation (BP) on a 0–10 scale using the same prompt template.

\section{User Study}
\label{app:user_study_supple}

\begin{figure}[t]
  \centering
  \includegraphics[width=\linewidth]{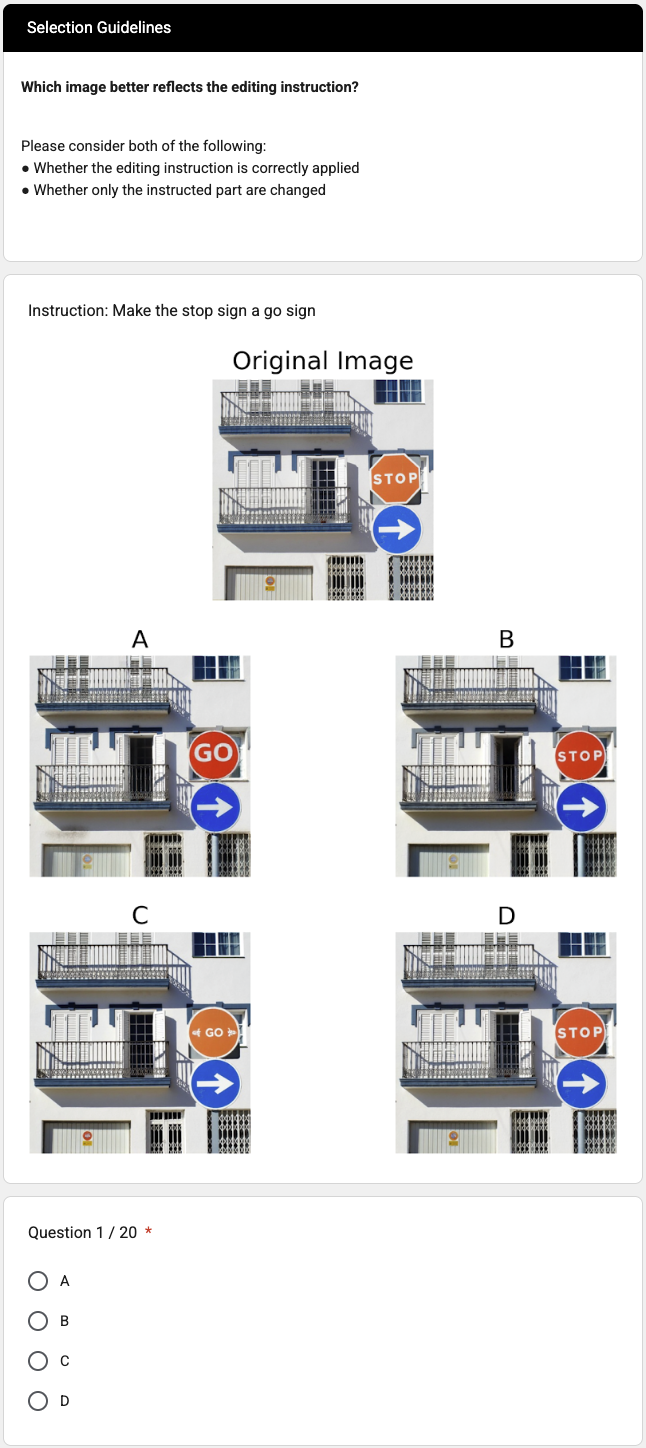}
  \caption{\textbf{User study interface.}
  Participants select the best result from four randomly ordered edits based on instruction alignment and preservation of unchanged content, given the source image and editing instruction.}
    \label{fig:user_study_interface}
\end{figure}

We recruit 70 participants, each evaluating 20 editing instances, yielding 1,400 responses. As shown in Fig.~\ref{fig:user_study_interface}, four edited results are presented anonymously in randomized order for each instance. The ordering is independently randomized across instances to minimize positional bias. The evaluation protocol and quantitative results are presented in Sec.~\ref{sec:user_study} of the main paper.

\begin{algorithm}[h]
\caption{Coupling-Controlled Candidate Construction}
\label{alg:candidate}
\begin{algorithmic}[1]
\Function{ConstructCandidate}{$t_i$}
    \State $N_{t_i}\sim\mathcal N(0,I)$
    \State $X_{t_i}^{\mathrm{src}}\leftarrow
    (1-t_i)X^{\mathrm{src}}+t_iN_{t_i}$
    \State $X_{t_i}^{\mathrm{tar}}\leftarrow
    X_{t_i}^{\mathrm{edit}}+
    X_{t_i}^{\mathrm{src}}-X^{\mathrm{src}}$
    \State $V_{t_i}^{\mathrm{cpl}}\leftarrow
    \mathcal D_{t_i}(X_{t_i}^{\mathrm{tar}},
    X_{t_i}^{\mathrm{src}})$
    \State $\mathcal V_{t_i}\leftarrow
    \{V_{t_i}^{\mathrm{cpl}}\}$
    
    \State $\tilde X_{t_i}^{\mathrm{src}}, \tilde X_{t_i}^{\mathrm{tar}} \leftarrow \varnothing,\varnothing$
    \If{$i\neq n_{\max}$}
        \State $\tilde X_{t_i}^{\mathrm{src}} \leftarrow \mathcal{P}_{t_{i+1}\to t_i}(X_{t_{i+1}}^{\mathrm{src}}, T_{\mathrm{src}})$
        \State $\tilde X_{t_i}^{\mathrm{tar}} \leftarrow \mathcal{P}_{t_{i+1}\to t_i}(X_{t_{i+1}}^{\mathrm{tar}}, T_{\mathrm{tar}})$
        \State $V_{t_i}^{\mathrm{dec}}\leftarrow
        \mathcal D_{t_i}(\tilde X_{t_i}^{\mathrm{tar}},
        \tilde X_{t_i}^{\mathrm{src}})$
        \State $V_{t_i}^{\mathrm{ext}}\leftarrow
        2V_{t_i}^{\mathrm{dec}}-V_{t_i}^{\mathrm{cpl}}$
        \State $\mathcal V_{t_i}\leftarrow
        \mathcal V_{t_i}\cup
        \{V_{t_i}^{\mathrm{dec}},V_{t_i}^{\mathrm{ext}}\}$
    \EndIf

    \If{$\bar X_{t_i}^{\mathrm{src}}\neq\varnothing$}
        \State $V_{t_i}^{\mathrm{pdec}}\leftarrow
        \mathcal D_{t_i}(\bar X_{t_i}^{\mathrm{tar}},
        \bar X_{t_i}^{\mathrm{src}})$
        \State $\mathcal V_{t_i}\leftarrow
        \mathcal V_{t_i}\cup\{V_{t_i}^{\mathrm{pdec}}\}$
    \EndIf
    \State \Return $\mathcal V_{t_i},
    \tilde X_{t_i}^{\mathrm{src}},
    \tilde X_{t_i}^{\mathrm{tar}}$
\EndFunction
\end{algorithmic}
\end{algorithm}

\begin{algorithm}[H]
\caption{Two-Stage Adaptive Masking}
\label{alg:masking}
\begin{algorithmic}[1]
\Function{AdaptiveMasking}{$V_{t}^*,s_{t_i},t_i,\tau$}
    \State $V_{\mathrm{V2C}}\leftarrow V_{t}^*$
    \If{$s_{t_i}\geq\tau$}
        \State $E_{t_i}(p)\leftarrow
        \|\hat X_{0|t_i}^{\mathrm{edit}}(V_{t}^*)(p)
        -X^{\mathrm{src}}(p)\|_2$
        \State $\eta_{t_i}\leftarrow
        \text{steepest drop in }
        \mathrm{hist}(\{E_{t_i}(p)\}_p)$
        \State $M_{t_i}(p)\leftarrow
        \mathbb I[E_{t_i}(p)>\eta_{t_i}]$
        \State $V_m\leftarrow M_{t_i}\odot V_{t}^*$
        \State $s_m\leftarrow
        \mathrm{ILS}(X^{\mathrm{src}},
        \hat X_{0|t_i}^{\mathrm{edit}}(V_m),
        T_{\mathrm{instr}})$
        \If{$s_m\geq s_{t_i}$}
            \State $V_{\mathrm{V2C}}\leftarrow V_m$
        \EndIf
    \EndIf
    \State \Return $V_{\mathrm{V2C}}$
\EndFunction
\end{algorithmic}
\end{algorithm}

\begin{figure*}[t]
  \centering
  \includegraphics[width=\linewidth]{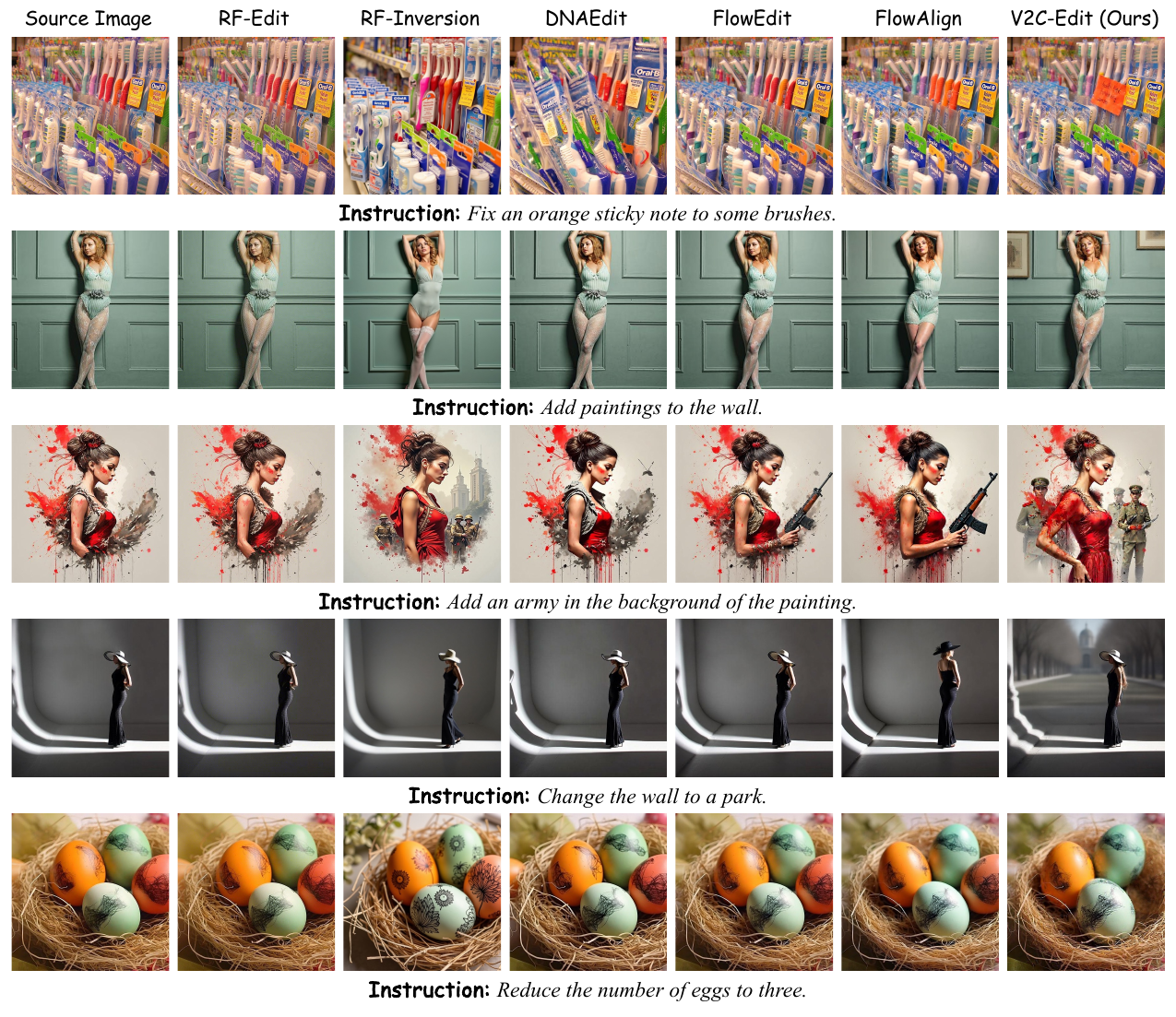}
    \caption{Additional qualitative results on FLUX.}
  \label{fig:qualitative_flux}
\end{figure*}

\begin{figure*}[t]
  \centering
  \includegraphics[width=\linewidth]{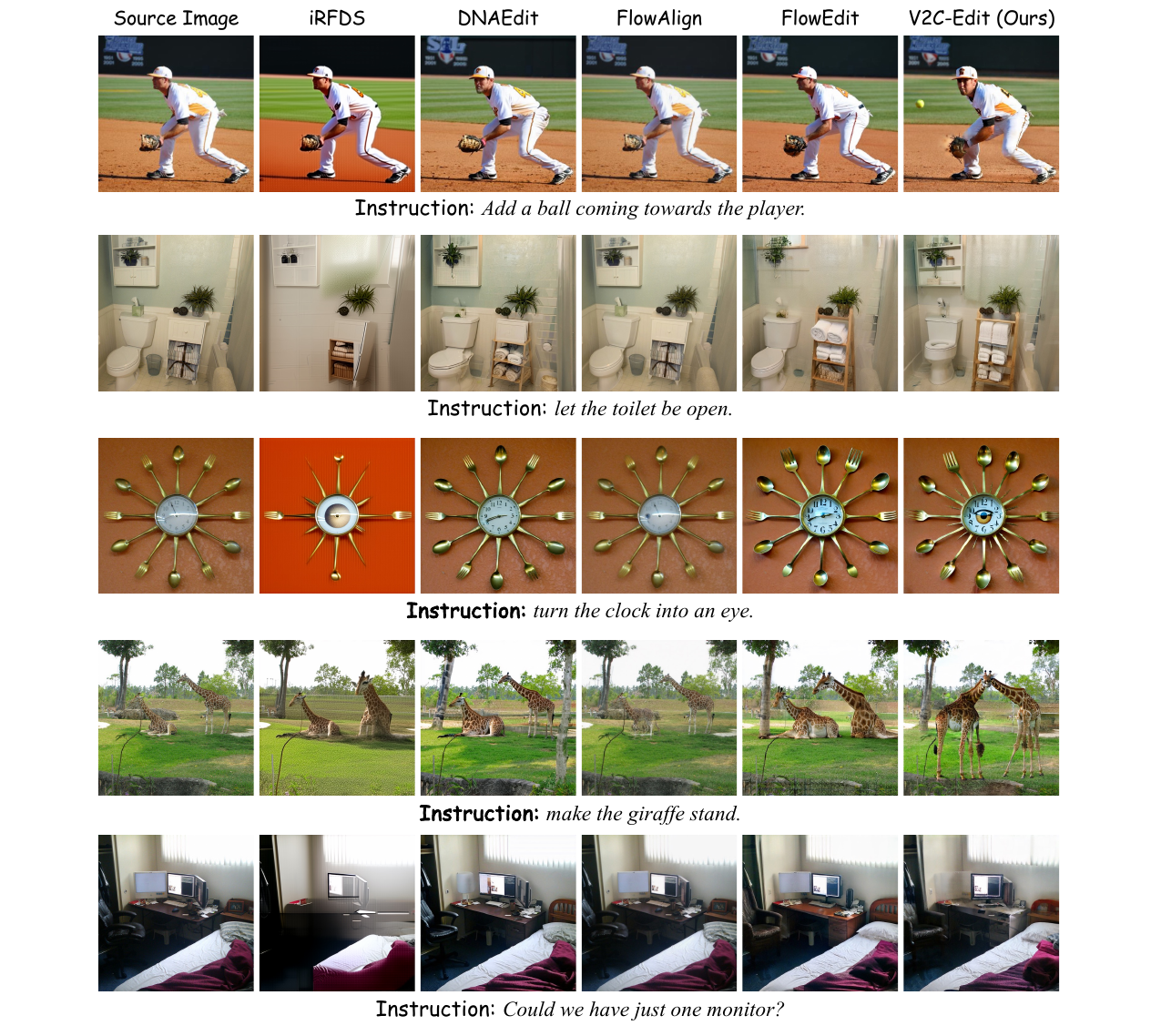}
  \caption{Additional qualitative results on SD3.}
  \label{fig:qualitative_SD3}
\end{figure*}

\clearpage
\onecolumn
\begin{promptbox}[VLM Query Templates for ILS Computation]
### Intended Transformation Query
# Input: source image, instruction
I will give you an editing instruction: '{instruction}'.
Combined with this image, give me a complete and detailed editing instruction in 20 words.

### Realized Transformation Query
# Input: source image, edited image 
Guess the image editing instruction from the first image to the second image in 20 words.
\end{promptbox}
\label{lst:ils-prompt}

\begin{promptbox}[Prompt Template for Human-Proxy Evaluation (GPT-4.1 and Gemini 3.6 Flash)] 
You are an evaluator for image editing. You will be given a pair of images before and after editing, along with an editing instruction.
Evaluate the editing result across three dimensions, each scored from 0 to 10:
1. **Instruction Alignment**: 
    How well does the edited image reflect the changes requested in the editing instruction? A successful edit should include all required changes and no extra unintended changes.
2. **Subject Preservation**: 
    How well are the unedited subjects/objects preserved?
    The structure, pose, orientation, and appearance of entities not targeted by the edit should remain consistent with the original.
3. **Background Preservation**: 
    How well is the background and surrounding context preserved?
    Regions outside the edited area should remain as close to the original as possible, with no unwanted alterations, color shifts, or distortions.

Be strict about changes between the two images. 
The second image should reflect only the minimum necessary changes to satisfy the editing instruction. Give the final response in a JSON format as such:
{
    "Instruction_Alignment": xx, 
    "Subject_Preservation": xx, 
    "Background_Preservation": xx
}
Do not output anything else.

### Input
- Editing instruction (text)
- Source image (before editing)
- Edited image (after editing)

### Output
{
    "Instruction_Alignment": xx, 
    "Subject_Preservation": xx, 
    "Background_Preservation": xx
}
\end{promptbox}
\label{lst:IA-prompt}

\twocolumn

\end{document}